\pdfoutput=1
\documentclass[a4paper,11pt]{article}

\usepackage{graphicx}  
\usepackage{times}
\usepackage{latexsym}
\usepackage[T1]{fontenc}
\usepackage[scaled=.8]{beramono}
\usepackage[scaled=.87]{helvet}
\usepackage{natbib}
\usepackage[margin=25mm]{geometry}
\usepackage{amsmath,amssymb}

\usepackage{subcaption}
\usepackage{sidecap}

\usepackage{xspace}

\usepackage{microtype}
\usepackage{multirow}

\usepackage{enumitem} 
\setitemize{noitemsep,topsep=0em} 
\setenumerate{noitemsep,leftmargin=2em,itemindent=13pt,topsep=0em}

\usepackage[usenames, dvipsnames]{color}
\definecolor{ltgray}{RGB}{215, 215, 215}
\definecolor{mypurple}{RGB}{112, 48, 160}

\newcommand{\svar}[1]
   {\setbox2=\hbox{$\scriptstyle #1$}\lower.2ex\vbox{\hrule
     \hbox{\vrule\kern1.25pt 
     \vbox{\kern1.25pt\box2\kern1.25pt}\kern1.25pt\vrule}\hrule}}
\newcommand{\SVAR}[1]
   {\setbox2=\hbox{$\scriptstyle #1$}\lower.2ex\vbox{\hrule
     \hbox{\bgroup\fboxsep=0pt\colorbox{ltgray}{\vrule\kern1.25pt 
     \vbox{\kern1.25pt\box2\kern1.25pt}\kern1.25pt\vrule}\egroup}\hrule}}

\newcommand{\citeposs}[2][]{\citeauthor{#2}'s (\citeyear[#1]{#2})}

\usepackage{hyperref}
\hypersetup{colorlinks=true,citecolor=black, linkcolor=., urlcolor=black}
\usepackage{cleveref}

\crefformat{part}{\S#2#1#3}
\crefformat{chapter}{\S#2#1#3}
\crefformat{section}{\S#2#1#3}
\crefformat{subsection}{\S#2#1#3}
\crefformat{subsubsection}{\S#2#1#3}
\crefformat{paragraph}{\P#2#1#3}
\crefformat{subparagraph}{\P#2#1#3}
\crefmultiformat{section}{\S#2#1#3}{ and~\S#2#1#3}{, \S#2#1#3}{, and~\S#2#1#3}
\crefmultiformat{subsection}{\S#2#1#3}{ and~\S#2#1#3}{, \S#2#1#3}{, and~\S#2#1#3}
\crefmultiformat{subsubsection}{\S#2#1#3}{ and~\S#2#1#3}{, \S#2#1#3}{, and~\S#2#1#3}
\crefmultiformat{paragraph}{\P\P#2#1#3}{ and~#2#1#3}{, #2#1#3}{, and~#2#1#3}
\crefmultiformat{subparagraph}{\P\P#2#1#3}{ and~#2#1#3}{, #2#1#3}{, and~#2#1#3}
\crefrangeformat{section}{\mbox{\S\S#3#1#4--#5#2#6}}
\crefrangeformat{subsection}{\mbox{\S\S#3#1#4--#5#2#6}}
\crefrangeformat{subsubsection}{\mbox{\S\S#3#1#4--#5#2#6}}
\crefrangeformat{paragraph}{\mbox{\P\P#3#1#4--#5#2#6}}
\crefrangeformat{subparagraph}{\mbox{\P\P#3#1#4--#5#2#6}}
\crefname{part}{Part}{Parts}
\Crefname{part}{Part}{Parts}
\crefname{chapter}{ch.}{ch.}
\Crefname{chapter}{Ch.}{Ch.}
\crefname{figure}{figure}{figures}
\crefname{subfigure}{figure}{figures}
\Crefname{subfigure}{Figure}{Figures}
\crefname{appsec}{appendix}{appendices}
\Crefname{appsec}{Appendix}{Appendices}
\crefname{algocf}{algorithm}{algorithms}
\Crefname{algocf}{Algorithm}{Algorithms}
\crefname{enums}{example}{examples}
\Crefname{enums}{Example}{Examples}
\crefname{enumsi}{example}{examples}
\Crefname{enumsi}{Example}{Examples}
\crefname{}{example}{examples} 
\Crefname{}{Example}{Examples}
\crefformat{enums}{(#2#1#3)}
\crefformat{enumsi}{(#2#1#3)}
\crefformat{}{(#2#1#3)}
\crefname{xnumi}{example}{examples} 
\crefname{xnumi}{example}{examples} 
\Crefname{xnumii}{Example}{Examples} 
\Crefname{xnumii}{Example}{Examples} 
\crefformat{xnumi}{(#2#1#3)} 
\crefformat{xnumii}{(#2#1#3)} 
\crefrangeformat{enums}{\mbox{(#3#1#4--#5#2#6)}}
\crefrangeformat{enumsi}{\mbox{(#3#1#4--#5#2#6)}}
\crefrangeformat{xnumi}{\mbox{(#3#1#4--#5#2#6)}} 
\crefrangeformat{xnumii}{\mbox{(#3#1#4--#5#2#6)}} 
\crefmultiformat{enumsi}{(#2#1#3}{, #2#1#3)}{, #2#1#3}{, #2#1#3)}
\crefmultiformat{xnumi}{(#2#1#3}{, #2#1#3)}{, #2#1#3}{, #2#1#3)} 
\crefmultiformat{xnumii}{(#2#1#3}{, #2#1#3)}{, #2#1#3}{, #2#1#3)} 
\crefrangemultiformat{enumsi}{(#3#1#4--#5#2#6}{, #3#1#4--#5#2#6)}{, #3#1#4--#5#2#6}{, #3#1#4--#5#2#6)}
\crefrangemultiformat{xnumi}{(#3#1#4--#5#2#6}{, #3#1#4--#5#2#6)}{, #3#1#4--#5#2#6}{, #3#1#4--#5#2#6)} 
\crefrangemultiformat{xnumii}{(#3#1#4--#5#2#6}{, #3#1#4--#5#2#6)}{, #3#1#4--#5#2#6}{, #3#1#4--#5#2#6)} 

\ifx\creflastconjunction\undefined%
\newcommand{\creflastconjunction}{, and\nobreakspace} 
\else%
\renewcommand{\creflastconjunction}{, and\nobreakspace} 
\fi%

\newcommand*{\Fullref}[1]{\hyperref[{#1}]{\Cref*{#1}: \nameref*{#1}}}
\newcommand*{\fullref}[1]{\hyperref[{#1}]{\cref*{#1}: \nameref{#1}}}
\newcommand{\fnref}[1]{fn.~\ref{#1}} 

\newcommand{\anonversion}[1]{}
\newcommand{\nonanonversion}[1]{#1}

\newcommand{\phrase}[2]{\multicolumn{#1}{c}{#2}}
\newcommand{\SEM}[1]{\textcolor{purple}{#1}}
\newcommand{\sem}[2]{\multicolumn{#1}{c}{\SEM{#2}}}
\newcommand{\semid}{\SEM{\textsc{id}}\xspace}
\newcommand{\ccgline}[2]{\multicolumn{#1}{c}{\hrulefill \raisebox{-.30ex}{\tiny \mbox{#2}}}}
\newcommand{\bs}{$\backslash$}
\newcommand{\fs}{$/$}

\newcommand{\combf}[1]{\ensuremath{\mathsf{#1}}} 
\newcommand{\combb}{\ensuremath{\combf{B}}}      
\newcommand{\combt}{\ensuremath{\combf{T}}}      

\newcommand{\combbx}{\mbox{\ensuremath{\combb_\times}}}

\newcommand{\combbb}{\ensuremath{\combf{B^2}}}

\newcommand{\combsa}{\ensuremath{\combf{R}}}
\newcommand{\combsb}{\ensuremath{\combf{RB}}}
\newcommand{\combsbb}{\ensuremath{\combf{RB^2}}}
\newcommand{\combconj}{\ensuremath{\combf{\&}}}

\newcommand{\qrel}{\textbf{:?}\xspace}

\usepackage{tikz}
\usetikzlibrary{shapes}

\title{An Improved Approach for Semantic Graph Composition with~CCG}
\date{}

\anonversion{\author{Example Author\\
       Affiliation\\
       \texttt{example@email.org}
  \and Someone Else\\
       Another Affiliation\\
       \texttt{another@email.org}
}}
\nonanonversion{\author{Austin Blodgett \quad Nathan Schneider\\
       Georgetown University, Department of Linguistics\\
       \{\texttt{ajb341}, \texttt{nathan.schneider}\}\texttt{@georgetown.edu}
}}

\begin{document}
\maketitle
\thispagestyle{empty}
\pagestyle{empty}

\begin{abstract}
This paper builds on previous work using Combinatory Categorial Grammar (CCG) to derive a transparent syntax-semantics interface for Abstract Meaning Representation (AMR) parsing. We define new semantics for the CCG combinators that is better suited to deriving AMR graphs. In particular, we define relation-wise alternatives for the application and composition combinators: these require that the two constituents being combined overlap in one AMR relation. We also provide a new semantics for type raising, which is necessary for certain constructions. Using these mechanisms, we suggest an analysis of eventive nouns, which present a challenge for deriving AMR graphs. Our theoretical analysis will facilitate future work on robust and transparent AMR parsing using CCG. 
\end{abstract}

\maketitle

\section{Introduction}

At the heart of semantic parsing are two goals: the disambiguation of linguistic forms that can have multiple meanings, and the normalization of morphological and syntactic variation.
Among many techniques for semantic parsing, one profitable direction exploits computational linguistic grammar formalisms that make explicit the correspondence between the linguistic form of a sentence and the semantics (e.g., broad-coverage logical forms, or database queries in a domain-specific query language).
In particular, English semantic parsers using Combinatory Categorial Grammar \citep[CCG;][]{steedman2000syntactic} have been quite successful thanks to the CCGBank resource \citep{hockenmaier-07,honnibal-10} and the broad-coverage statistical parsing models trained on it \citep[e.g.,][]{clark-04,lewis-16,clark-18}.

The CCG formalism assumes that all language-specific grammatical information is stored in a lexicon: each word in the lexicon is associated with a structured syntactic \textbf{category} and a semantic form, such that the compositional potentials of the category and the semantics are isomorphic. A small universal set of \textbf{combinators} are responsible for assembling constituents into a full syntactic derivation; each combinator operates on adjacent constituents with appropriate categories to produce a new constituent and its compositional semantics, subject to constraints. A full grammar thus allows well-formed sentences to be transduced into semantic structures.
The categories and combinators cooperate to license productive syntactic constructions like control and wh-questions, requiring the correct word order and producing the correct semantic dependencies. For example, consider the sentence ``Who did John seem to forget to invite to attend?'': the correct logical form---in propositional logic, something like  $\textit{seem}(\textit{forget}(\textit{John}_i, \textit{invite}(\textit{John}_i, \textit{who}_j, \textit{attend}(\textit{who}_j))))$---is nontrivial, requiring a precise account of several constructions that conspire to produce long-range dependencies.

Whereas CCG traditionally uses some version of lambda calculus for its semantics, there has also been initial work using CCG to build parsers for Abstract Meaning Representation \citep[AMR;][]{amr}, a standard with which a large ``sembank'' of English sentences\footnote{See \url{https://amr.isi.edu/download.html}} has been manually annotated.\footnote{As originally defined, AMR is English-specific. However, a companion annotation standard, corpus, and parsers exist for Chinese \citep{xue-14,li-16,wang-18}, and initial investigations have been made toward adapting AMR to several other languages \citep{xue-14,migueles-abraira-18,anchieta-18}.}
To date, dozens of publications\footnote{\url{https://nert-nlp.github.io/AMR-Bibliography/} is a categorized list of publications about or using AMR.} have used the corpus to train and evaluate semantic parsers---most using graph-based or transition-based parsing methods \citep[e.g.,][]{flanigan-14,wang2016camr,lyu-18} to transform the sentence string or syntactic parse into a semantic graph via a learned statistical model, without any explicit characterization of the syntax-semantics interface.
There is good reason to apply CCG to the AMR parsing task: 
apart from transparency of the syntax-semantics interface,
state-of-the-art AMR parsers are known to be weak at reentrancy \citep[e.g.,][]{lyu-18}, which presumably can be partially attributed to syntactic reentrancy in control constructions, for example. 
Prior work applying CCG to AMR parsing has begun to address this, but some of the important mechanisms that make CCG a linguistically powerful and robust theory have yet to be incorporated into these approaches.



In this paper, we build on a core insight of previous work \citep[e.g.,][]{artzi2015broad,beschke2018graph} that AMR fragments can be directly represented as the semantics of CCG lexical entries. With appropriate definitions of the lexical items and combinatorial rules of CCG, the compositionality of CCG gives a derivation of a full AMR ``for free''. In other words, AMR parsing can be reduced to CCG parsing (plus some additional semantic disambiguation and postprocessing). On a practical level, this should allow us to take advantage of existing CCG datasets and parsing methods for AMR parsing. 
In addition, explicitly storing AMR fragments in the CCG lexicon would provide a level of interpretability not seen in most statistical AMR parsers: the transparent syntax-semantics interface would decouple errors in the grammar from errors in the parsing model.


As a prerequisite for building a CCG-based AMR parser, or inducing a broad-coverage grammar (CCG lexicon) from data, we consider in this paper the formal mechanisms that would be necessary to derive AMRs with linguistic robustness.
In particular, we address a variety of challenging syntactic phenomena with respect to AMR, showing the semantic fragments, associated syntactic categories, and combinators that will facilitate parsing of constructions including control, wh-questions, relative clauses, case marking, nonconstituent coordination, eventive nouns, and light verbs. 
In so doing, we offer new semantics of combinators for semantic graphs beyond the proposals of previous work.

After an overview of related work (\cref{sec:rel-work}),\footnote{Due to space constraints, we assume the reader is familiar with the basics of both CCG and AMR.} we introduce our formalism for AMR graph semantics in CCG (\cref{sec:graph-sem}). \Cref{sec:ling-examples} gives example derivations for well-known linguistic phenomena including control, complex coordination, and eventive nouns. \Cref{sec:discussion} discusses some implications of our approach.

\section{Related Work}\label{sec:rel-work}


AMR formalizes sentence meaning via a graph structure. 
The AMR for an English sentence is a directed acyclic graph that abstracts away from morphological and syntactic details such as word order, voice, definiteness, and morphology, focusing instead on lexical semantic predicates, roles, and relations. 
Semantic predicate-argument structures are based on the PropBank frame lexicon \citep{kingsbury2002treebank} and its frame-specific core argument roles (named \textit{ARG0}, \textit{ARG1}, etc.).
AMR supplements these with its own inventory of noncore relations like \SEM{:time} and \SEM{:purpose}, and some specialized frames for the semantics of comparison, for example.
Named entities are typed and linked to Wikipedia pages; dates and other values are normalized. Edges in the graph correspond to roles\slash relations, and nodes to predicate or non-predicate ``concepts'', which are lemmatized. Reentrancy is used for within-sentence coreference.

A limited amount of prior research has combined CCG and AMR.  \Citet{artzi2015broad} and \citet{misra2016neural} develop an AMR parser using CCG by reformulating AMR graphs as logical forms in lambda calculus. 
We opt here for an approach similar to that of \citet{beschke2018graph}, where AMR subgraphs with free variables are treated as the semantics in the CCG lexicon.
This requires definitions of the combinators that operate directly on AMR subgraphs rather than lambda calculus expressions.

\Citet{beschke2018graph} situate their formalization within the literature on graph grammars. 
They formulate their approach in terms of the HR algebra \citep{courcelle-12}, which \citet{koller2015semantic} had applied to AMR graphs (but not with CCG).
In this formalism, graph fragments called s-graphs are assembled to derive full graphs. S-graphs are equivalent to the AMR subgraphs described in this paper.

In particular, \citeauthor{beschke2018graph} define the semantics of CCG combinators in terms of HR-algebraic operations on s-graphs.
They discuss a small set of combinators from \citet{lewis2014ccg} that includes forward and backward application and forward, backward, crossed, and generalized variants of composition.
We introduce equivalent semantics for application and composition (\cref{sec:FA}), avoiding the conceptually heavy notation and formalism from the HR algebra. 
They also specify Conjunction and Identity combinators, which we adapt slightly to suit our needs, and a Punctuation combinator.
More significantly, they treat unary operators such as type raising to have no effect on the semantics, whereas we will take another route for type raising (\cref{sec:tr}), and will introduce new, \emph{relation-wise} versions of application and composition (\cref{sec:sa}).
Finally, whereas \citeauthor{beschke2018graph} devote most of their paper to a lexicon induction algorithm and experiments, we focus on the linguistic motivation for our definition of the combinators, and leave the development of suitable lexicon induction techniques to future work.


A related graph formalism called \emph{hyperedge replacement grammar} is also used in the AMR parsing literature \citep{jones2012semantics,chiang2013parsing,peng2015synchronous,peng2016uofr,bjorklund2016between,groschwitz2018amr}. Hyperedge replacement grammars \citep{rozenberg1997handbook} are a formal way of combining subgraphs to derive a larger graph, based on an extension of Context Free Grammars to graphs instead of strings. Readers may assume that the graph formalism described in this paper is a simplified hyperedge replacement grammar which only allows hyperedges of rank 1.

\section{Graph Semantics}\label{sec:graph-sem}
AMR is designed to represent semantics at the sentence level. For CCG lexical entries and combinators to parse AMR semantics, we need to formalize how AMR subgraphs can represent the semantics of individual words, and how combinators combine subgraphs to derive a full AMR. 
This section will formalize AMR subgraph semantics and CCG combinators for \textit{function application}, \textit{composition}, and \textit{type raising}. Additionally, we propose new \textit{relation-wise} variants of application and composition which are unique to graph semantics. 

\begin{SCfigure}[2][t]\centering
    \caption{Basic shape of AMR subgraph: Free variables (square, blue) are represented with \textit{x, y, z}, etc. AMR nodes (round, red) are represented with \textit{a, b, c}, etc. Dots indicate that part of the graph may be present or not.}
    \resizebox{!}{2.3cm}{
\begin{tikzpicture}[
white/.style={rectangle, minimum size=7mm},
blue/.style={rectangle, draw=blue!60, fill=blue!5, very thick, minimum size=7mm},
purple/.style={ellipse, draw=purple!60, fill=purple!5, very thick, minimum size=8mm},
purple-dot/.style={ellipse, draw=purple!60, fill=purple!5, very thick, dotted, minimum size=7mm},
blue-dot/.style={rectangle, draw=blue!60, fill=blue!5, very thick, dotted, minimum size=7mm},
]
    \begin{scope}[scale=0.7, transform shape]
    	\node[white](a) at (6, 1.35) {X~~~:};
    	\node[purple](p) at (10.0, 3) {p};
    	\node[purple-dot](x1) at (6.666666666666667,0.0) {a};
    	\node[purple-dot](x2) at (8.888888888888889,0.0) {b};
    	\node[white](dots) at (10,0.0) {$\dots$};
    	\node[blue-dot](x4) at (11.11111111111111,0.0) {x};
    	\node[blue-dot](y) at (13.333333333333334,0.0) {y};
    	\draw[->, thick, dotted] (p.south) -- (x1.north) node[midway, above, sloped] {:rel1};
    	\draw[->, thick, dotted] (p.south) -- (x2.north) node[midway, above, sloped] {:rel2};
    	\draw[->, thick, dotted] (p.south) -- (x4.north) node[midway, above, sloped] {:rel3};
    	\draw[->, thick, dotted] (p.south) -- (y.north) node[midway, above, sloped] {:rel4};
	\end{scope}
\end{tikzpicture}
}
\label{basic}
\end{SCfigure}
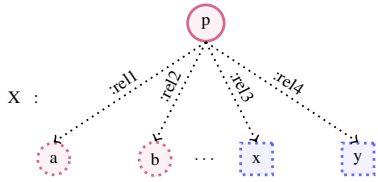

Each AMR subgraph contains nodes and edges from the resulting AMR as well as some nodes which correspond to free variables. The basic shape of an AMR subgraph appears in \cref{basic}.
Formally, an AMR subgraph is a tuple $\langle G, R, FV\rangle$, where
 $G$ is a connected, labeled, directed acyclic graph;
 $R$ is the root node in $G$;
 and $FV$ is an ordered list of the nodes of $G$ which are free and must be substituted by the end of the derivation.
Though not shown in \cref{basic}, the root of an AMR subgraph may be a free variable. Intuitively, a subgraph with at least one free variable corresponds to a function, and a subgraph with no free variables corresponds to a constant.

\textbf{Textual notation.} 
Taking inspiration from the PENMAN notation used for AMR, we use the notation
$\left(a~\text{:rel1}~(\svar{2}~\text{:rel2}~\svar{1})\right)$
to denote an AMR subgraph rooted at a constant $a$, 
with a \SEM{:rel1} edge to a free variable, \svar{2}, which in turn has a child free variable, \svar{1}.

\Cref{all-combinators} shows the formulation of graph semantics for all the combinators described below. The formulas are schematic with attention paid to the resulting order of free variables, which semantically distinguishes application from composition. Another combinator in CCG, crossing composition, has the same semantics as regular composition. Semantics for the substitution combinator is left to future work.

\begin{table}[t]
    \centering\footnotesize 
    \begin{tabular}{@{}c@{~}|@{~}c@{~}|@{~}c@{~}|@{~}c@{~}|@{~}c@{}}
    \textit{combinator} & \textit{function (left/right)} & \textit{arg.~(right/left)} & \textit{result} & \textit{FV ordering} \\
    \hline
    \multicolumn{5}{|@{~}c@{~}|}{\textbf{Binary}}\\
    \hline
         \textbf{\textsf{Application}}
         & $\dots_1$ $\svar{1}$ $\dots_2$ 
         & a~$\dots_3$
         & $\dots_1$ a~$\dots_2\dots_3$  & $\svar{2},\dots,\SVAR{1},\dots$ \\
         \textbf{\textsf{Composition}} (\combb, \combbb)
         & $\dots_1$ $\svar{1}$ $\dots_2$ 
         & a~$\dots_3$
         & $\dots_1$ a~$\dots_2\dots_3$  & $\SVAR{1},\dots,\svar{2},\dots$ \\
          \hline
         \textbf{\textsf{Relation-wise Application}} (\combsa)
         & $\dots_1$ $\svar{1}$~~:rel$_x$~~b~$\dots_2$
         & ~a~~:rel$_x$~~$\SVAR{1}$~$\dots_3$
         & $\dots_1$ a~~:rel$_x$~~b~$\dots_2\dots_3$ & $\svar{2},\dots,\SVAR{2},\dots$ \\
         \textbf{\textsf{Relation-wise Composition}} (\combsb)
         & $\dots_1$ $\svar{1}$~~:rel$_x$~~b~$\dots_2$
         & ~a~~:rel$_x$~~$\SVAR{2}$~$\dots_3$
         & $\dots_1$ a~~:rel$_x$~~b~$\dots_2\dots_3$ & $\SVAR{1},\SVAR{3},\dots,\svar{2},\dots$ \\
         \textbf{\textsf{\dots Second-order}} (\combsbb)
         & $\dots_1$ $\svar{1}$~~:rel$_x$~~b~$\dots_2$
         & ~a~~:rel$_x$~~$\SVAR{3}$~$\dots_3$
         & $\dots_1$ a~~:rel$_x$~~b~$\dots_2\dots_3$ & $\SVAR{1},\SVAR{2},\SVAR{4},\dots,\svar{2},\dots$ \\
         \hline
    \multicolumn{5}{|@{~}c@{~}|}{\textbf{Unary}}\\
        \hline
         \textbf{\textsf{Type Raising}} (\combt) & \multicolumn{2}{@{~}c@{~}|}{a $\dots_1$}& $\svar{1}$~~\qrel~~a $\dots_1$ & $\svar{1},\SVAR{1},\dots$ \\
         \hline
    \multicolumn{5}{|@{~}c@{~}|}{\textbf{N-ary} ($\leq$1 FV per operand)}\\
        \hline
         \textbf{\textsf{Conjunction}} (\combconj) & x & \footnotesize  a $\dots_1$, b $\dots_2$, $\dots$  & \footnotesize x :op1 a $\dots_1$ :op2 b $\dots_2$ $\dots$ & $\svar{1}$ \\
    \end{tabular}
    \caption{Formal semantic rules for AMR combinators. Boxed numbers stand for free variables (FVs) in the semantics of each of the constituents being combined: $\svar{1}$ stands for the lowest indexed FV in the function (head) constituent, and $\SVAR{1}$ for the lowest indexed FV in the argument constituent, if any.
    Ellipses $\dots_n$ denote optional dominating structure (if preceding) and optional dominated structure (if following). 
    Any FVs in these optional structures are preserved in the result, in the order given in the last column. 
    For relation-wise combinators, the function constituent's relation may also be \qrel.
    Crossing composition (\combbx) and its variants behave semantically like their non-crossing counterparts. Not shown: exceptions to application and composition for the identity function (\semid), discussed in \cref{sec:id}.}
    \label{all-combinators}
\end{table}

\subsection{Syntax-Semantics Isomorphism}\label{sec:iso}

A core property of CCG is that it provides transparency in the syntax-semantics interface: both syntactic categories and semantic forms are defined as functions permitting a compositional derivation of the sentence. 
The syntactic category determines which constituents may be constructed and in what word order.
In the semantics, the word order (direction of the slashes) is irrelevant, but the functional structure---the arity and the order in which arguments are to be applied---must match in order for the semantics to remain well-formed as the sentence is derived based on the syntactic categories and combinatorial rules.

In other words, the functional structure of the category must be isomorphic to the functional structure of the semantics. 
For example, a hypothetical CCG category V\bs W\fs X\fs (Y\fs Z) would naturally correspond to a ternary function whose first argument, Y\fs Z, is itself a unary function.

This brings us to the following principle:

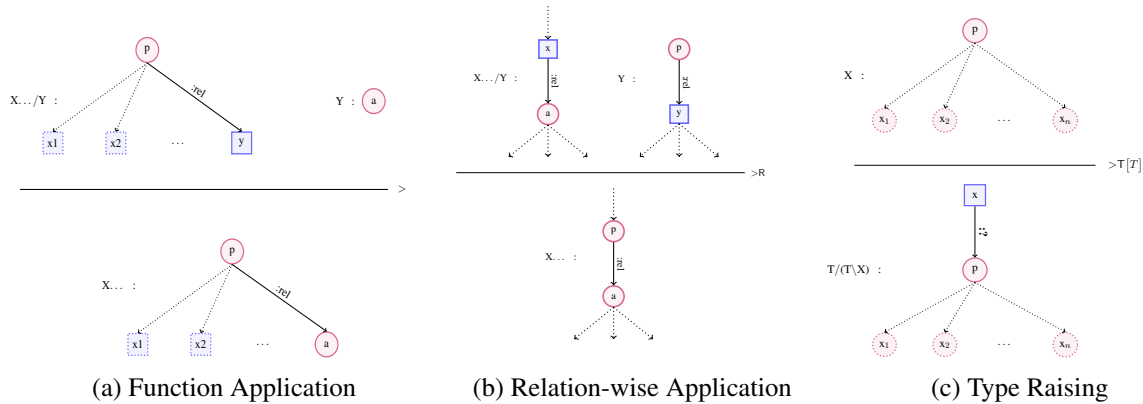
\begin{figure}[bt]
\centering
\begin{subfigure}[t]{5.8cm}
\resizebox{5.4cm}{4.25cm}{
\begin{tikzpicture}[
white/.style={rectangle, minimum size=7mm},
blue/.style={rectangle, draw=blue!60, fill=blue!5, very thick, minimum size=7mm},
purple/.style={ellipse, draw=purple!60, fill=purple!5, very thick, minimum size=8mm},
blue-dot/.style={rectangle, draw=blue!60, fill=blue!5, very thick, dotted, minimum size=7mm},
purple-dot/.style={ellipse, draw=purple!60, fill=purple!5, very thick, dotted, minimum size=7mm},
]
    	\node[white](a) at (6, 2.35) {X$\dots$\fs Y~~~:};
    	\node[white](b) at (17, 2.35) {Y~~~:};
    	\node[purple](p) at (10.0, 4) {p};
    	\node[purple](y2) at (18, 2.35) {a};
    	\node[blue-dot](x1) at (6.666666666666667,1.0) {x1};
    	\node[blue-dot](x2) at (8.888888888888889,1.0) {x2};
    	\node[white](x3) at (11.11111111111111,1.0) {$\dots$};
    	\node[blue](y) at (13.333333333333334,1.0) {y};
    	\draw[->, thick, dotted] (p.south) -- (x1.north) node[midway, above, sloped] {};
    	\draw[->, thick, dotted] (p.south) -- (x2.north) node[midway, above, sloped] {};
    	\draw[->, thick] (p.south) -- (y.north) node[midway, above, sloped] {:rel};
    	
        \draw (5.5,-0.5) -- (18.5,-0.5);
        \node[white](F) at (19,-0.5){$>$};
        
    	\node[white](a) at (9, -3.60) {X$\dots$~~~:};
    	\node[purple](p) at (13.0, -2.5) {p};
    	\node[blue-dot](x1) at (9.666666666666667,-5.5) {x1};
    	\node[blue-dot](x2) at (11.888888888888889,-5.5) {x2};
    	\node[white](x3) at (14.11111111111111,-5.5) {$\dots$};
    	\node[purple](y) at (16.333333333333334,-5.5) {a};
    	\draw[->, thick, dotted] (p.south) -- (x1.north) node[midway, above, sloped] {};
    	\draw[->, thick, dotted] (p.south) -- (x2.north) node[midway, above, sloped] {};
    	\draw[->, thick] (p.south) -- (y.north) node[midway, above, sloped] {:rel};
\end{tikzpicture}
}
\caption{Function Application}
\label{FA}
\end{subfigure}
\begin{subfigure}[t]{4.75cm}
\resizebox{!}{4.9cm}{
\begin{tikzpicture}[
white/.style={rectangle, minimum size=7mm},
blue/.style={rectangle, draw=blue!60, fill=blue!5, very thick, minimum size=7mm},
purple/.style={ellipse, draw=purple!60, fill=purple!5, very thick, minimum size=8mm},
purple-dot/.style={rectangle, draw=purple!60, fill=purple!5, very thick, dotted, minimum size=7mm},
]
    \begin{scope}[scale=0.7, transform shape]
    	\node[white](a) at (8, 0.35) {X$\dots$\fs Y~~~:};
    	\node[white](b1) at (10.0, 3.5) {};
    	\node[blue](p) at (10.0, 1.5) {x};
    	\node[purple](y) at (10, -1) {a};
    	\node[white](b2) at (8.5, -3) {};
    	\node[white](b3) at (10, -3) {};
    	\node[white](b4) at (11.5, -3) {};
    	\draw[->, thick] (p.south) -- (y.north) node[midway, above, sloped] {:rel};
    	\draw[->, thick, dotted] (b1.south) -- (p.north) node[midway, above, sloped] {};
    	\draw[->, thick, dotted] (y.south) -- (b2.north) 
    	node[midway, above, sloped] {};
    	\draw[->, thick, dotted] (y.south) -- (b3.north) node[midway, above, sloped] {};
    	\draw[->, thick, dotted] (y.south) -- (b4.north) node[midway, above, sloped] {};
    	
    	\node[white](a2) at (13, 0.35) {Y~~~:};
    	\node[purple](p2) at (15, 1.5) {p};
    	\node[blue](y2) at (15.0,-1) {y};
    	\node[white](b6) at (13.5, -3) {};
    	\node[white](b7) at (15, -3) {};
    	\node[white](b8) at (16.5, -3) {};
    	\draw[->, thick] (p2.south) -- (y2.north) node[midway, above, sloped] {:rel};
    	\draw[->, thick, dotted] (y2.south) -- (b6.north) 
    	node[midway, above, sloped] {};
    	\draw[->, thick, dotted] (y2.south) -- (b7.north) node[midway, above, sloped] {};
    	\draw[->, thick, dotted] (y2.south) -- (b8.north)
    	node[midway, above, sloped] {};
    	
        \draw (6.5,-3.25) -- (17.5,-3.25);
        \node[white](F) at (18,-3.25){$>$\combsa};
        
        \node[white](b9) at (12.5, -3.5) {};
        \node[white](b10) at (11, -10) {};
        \node[white](b11) at (12.5, -10) {};
        \node[white](b12) at (14, -10) {};
        \node[white](a3) at (10.5, -6.5) {X$\dots$~~~:};
    	\node[purple](p3) at (12.5, -5.5) {p};
    	\node[purple](y3) at (12.5,-8) {a};
    	\draw[->, thick] (p3.south) -- (y3.north) node[midway, above, sloped] {:rel};
    	\draw[->, thick, dotted] (b9.south) -- (p3.north) 
    	node[midway, above, sloped] {};
    	\draw[->, thick, dotted] (y3.south) -- (b10.north) 
    	node[midway, above, sloped] {};
    	\draw[->, thick, dotted] (y3.south) -- (b11.north) node[midway, above, sloped] {};
    	\draw[->, thick, dotted] (y3.south) -- (b12.north)
    	node[midway, above, sloped] {};
    \end{scope}
\end{tikzpicture}
}
\caption{Relation-wise Application}
\label{SA}
\end{subfigure}
\begin{subfigure}[t]{5.25cm}
\resizebox{!}{4.5cm}{
\begin{tikzpicture}[
white/.style={rectangle, minimum size=7mm},
purple/.style={ellipse, draw=purple!60, fill=purple!5, very thick, minimum size=8mm},
blue/.style={rectangle, draw=blue!60, fill=blue!5, very thick, minimum size=7mm},
purple-dot/.style={circle, draw=purple!60, fill=purple!5, very thick, dotted, minimum size=7mm},
]
    
        \node[white](a) at (-4, 3) {X~~~:};
    	\node[purple](p) at (0, 4.5) {p};
    	\node[purple-dot](x1) at (-3, 1.5) {x$_1$};
    	\node[purple-dot](x2) at (-1 , 1.5) {x$_2$};
    	\node[white](x3) at (1,1.5) {$\dots$};
    	\node[purple-dot](x4) at (3,1.5) {x$_n$};
    	\draw[->, thick, dotted] (p.south) -- (x1.north) node[midway, above, sloped] {};
    	\draw[->, thick, dotted] (p.south) -- (x2.north) node[midway, above, sloped] {};
    	\draw[->, thick, dotted] (p.south) -- (x4.north) node[midway, above, sloped] {};
    
    	\draw (-4,0) -- (4,0);
    	\node[white](f) at (5, 0) {$>$\combt$\big[T\big]$};
    	
    	\node[blue](x) at (0, -1) {x};
    	\node[white](a2) at (-4, -3.5) {T\fs (T\bs X)~~~:};
    	\node[purple](p2) at (0, -3.5) {p};
    	\node[purple-dot](x5) at (-3, -6) {x$_1$};
    	\node[purple-dot](x6) at (-1 ,-6) {x$_2$};
    	\node[white](x7) at (1,-6) {$\dots$};
    	\node[purple-dot](x8) at (3,-6) {x$_n$};
    	\draw[->, thick, dotted] (p2.south) -- (x5.north) node[midway, above, sloped] {};
    	\draw[->, thick, dotted] (p2.south) -- (x6.north) node[midway, above, sloped] {};
    	\draw[->, thick, dotted] (p2.south) -- (x8.north) node[midway, above, sloped] {};
    	\draw[->, thick] (x.south) -- (p2.north) node[midway, above, sloped] {\qrel};
    	
    	
\end{tikzpicture}
}
\caption{Type Raising}
\label{TR}
\end{subfigure}
\caption{Combinators illustrated in terms of semantic graph structure. The semantics of composition differs from application only in ordering of free variables (not shown).}
\end{figure}

\paragraph{\textit{Principle of Functional Isomorphism.}}
The semantics of a word or constituent cannot have higher arity than the CCG category calls for, and every functional category must take at least one semantic argument.
For instance, a word or constituent with category PP\fs NP must have exactly 1 semantic argument; and the VP adjunct category (S\bs NP)\bs(S\bs NP) a.k.a.~S\bs NP\bs(S\bs NP) can be interpreted as having 1 or 2 semantic arguments.

Without proving it formally, we remark that
this helps ensure that syntactic well-formedness according to the categories will guarantee semantic well-formedness, with no attempt to apply something that is not expecting any arguments, and no free variables remaining in the semantics at the end of a sentence derivation.
(An edge case where this guarantee might not hold is noted in \fnref{fn:underspecified}.)

\subsection{Function Application and Composition}\label{sec:FA}\label{sec:B}

In \textbf{Function Application} of AMR subgraphs, a free variable (blue) can be filled by the root of another AMR subgraph. The case of right function application is shown in \cref{FA}. Function application can only substitute the first free variable in $FV$ corresponding to the rightmost syntactic argument. 

While application and composition always differ syntactically, from a graph semantics point of view, composition turns out to be the same as function application, where the root of one subgraph is substituted for a free variable in another subgraph. The difference between application and composition is captured in the resulting order of free variables. In the case of composition, the argument's free variables are placed first on the free variable stack followed by the function's free variables. This allows free variables in the AMR subgraph to consistently match syntactic arguments in the CCG category. This is a difference between composition in this work and in \citeposs{beschke2018graph} work, where the semantics of application and composition is the same.

\subsection{Relation-wise Application and Composition}\label{sec:sa}\label{sec:sb}

When deriving a constituent, there are situations where it is desirable to have a semantic edge that is shared between the two constituents being combined. 
For example, we specify the following lexical entry for the control verb ``decide'', indexing arguments in the category with subscripts for clarity: S$_{\textit{b}}$\bs NP$_2$\fs (S$_{\textit{to}}$\bs NP)$_1$~: \SEM{decide-01 :ARG0 $\svar{2}$ ~:ARG1 ($\svar{1}$ \textbf{:ARG0} $\svar{2}$)}.
Unlike a simple verb, ``decide'' selects for an embedded clause and controls its subject, coindexing it with the matrix subject. This is indicated in the semantics with the bolded \SEM{:ARG0} edge, which needs to unify with the \SEM{:ARG0} edge of the embedded predicate. 
Thus the constituent ``you decide to eat yesterday'' in \cref{wh-control} 
is formed by merging the \SEM{:ARG0} edge expected by ``decide'' and the \SEM{:ARG0} edge expected by ``eat'' so that they may later be filled by the same node, \SEM{you}.
Note that the number of semantic free variables respects the functional structure of the category (\cref{sec:iso}).
%
To facilitate this, we define novel \textbf{relation-wise} variants of the  application and composition combinators that expect an edge in common (call it the \textbf{shared edge}).
Apart from control, relation-wise combinators are also useful for derivations with type raising and various interesting syntactic constructions.

The schematic graph structures serving as inputs and outputs for relation-wise combinators are shown in \cref{SA}, and the full definition is given in \cref{all-combinators}. Notably, the function constituent has its lowest-indexed free variable at the \emph{source} of the shared edge, 
and the argument constituent has a free variable at the \emph{target} of the shared edge (the variable's index depending on the kind of application or composition). 
In the result, each free variable unifies with the node or variable at the same side of the edge in the other constituent. 
Other material attached to the shared edge in either constituent will be preserved in the result.

The regular vs.~relation-wise distinction applies only to the \emph{semantics}; syntactically, relation-wise application (composition) is just like regular application (composition).
During parsing, relation-wise combinators apply if and only if the two constituents being combined share a common relation with the appropriate free variables; otherwise, the non--relation-wise version of the combinator is used.

\textbf{Relation-wise Composition} (\combsb) differs from \textbf{Relation-wise Application} (\combsa) in the index of the argument's free variable being unified and in the resulting order of free variables. 
Just as regular composition can be used to adjust the order that constituents are normally combined and ``save an argument for later'', relation-wise composition 
does this with respect to a common edge.
Examples of both relation-wise and non--relation-wise composition appear in \cref{wh-control}. 

\subsection{Type Raising}\label{sec:tr}

In CCG, \textbf{Type Raising} (\combt) converts an argument into a function.
For example, the nominative case of the pronoun ``I'' can be coded in the syntactic category by making it a function that expects a verb phrase on the right and returns a sentence, thus preventing ``I'' from serving as an object.
For our framework to support type raising, we need an appropriate semantic conversion that respects the functional structure of the category---thus, the type-raised semantics must take an argument. 
However, as type raising can be applied to different types of arguments, we do not know a priori which relation label to produce.
Therefore, we introduce the notion of an \textbf{underspecified edge}, denoted \qrel. The type-raised structure has a free variable at the source of the underspecified edge, with the original subgraph at the target, as shown in \cref{TR}.
For example, see ``John'' and ``Mary'' in \cref{coordination}, where type raising is necessary to support subject+verb constituents for coordination.
The type-raised constituent must eventually be the input to a relation-wise combinator, which will specify the label on the edge.



Note that in this strategy of representing type raising, the isomorphism between functions in semantics and syntactic category is maintained. This fits with CCG's philosophy of a transparent syntax-semantics interface (\cref{sec:iso}). By contrast, \citeposs{beschke2018graph} strategy was to leave the result of type raising semantically unchanged, creating a mismatch between the syntax and the semantics. 



\section{Linguistic Examples}\label{sec:ling-examples}
This section explains the use of the combinators discussed in \cref{sec:graph-sem} for particular linguistic constructions.

\begin{figure}[t]
\centering
\begin{subfigure}[b]{0.35\textwidth}
\resizebox{!}{1.75cm}{
\begin{tikzpicture}[
blue/.style={rectangle, draw=blue!60, fill=blue!5, very thick, minimum size=7mm},
purple/.style={ellipse, draw=purple!60, fill=purple!5, very thick, minimum size=8mm},
white/.style={rectangle, minimum size=7mm},
]
    
    \begin{scope}[scale=0.7, transform shape]
    	\node[purple](e) at (10.0,2.5) {r/read-01};
    	\node[blue](x) at (7,0.0) {$x_2$};
    	\node[blue](y) at (13,0.0) {$x_1$};
    	\node[white](l) at (5.0,1.25) {S\bs NP$_2$\fs NP$_1$~~~:};
    	\draw[->, thick] (e.south) -- (x.north) node[midway, above, sloped] {:ARG0};
    	\draw[->, thick] (e.south) -- (y.north) node[midway, above, sloped] {:ARG1};
	\end{scope}
\end{tikzpicture}}
\caption{\raggedright\footnotesize``read''; \SEM{(r/read-01 :ARG0~$\svar{2}$ :ARG1~$\svar{1}$)}}
\label{read}
\end{subfigure}
\begin{subfigure}[b]{0.27\textwidth}\centering
\resizebox{!}{1.8cm}{
\begin{tikzpicture}[
blue/.style={rectangle, draw=blue!60, fill=blue!5, very thick, minimum size=7mm},
purple/.style={ellipse, draw=purple!60, fill=purple!5, very thick, minimum size=8mm},
white/.style={rectangle, minimum size=7mm},
]
    \begin{scope}[scale=0.7, transform shape]
    	\node[blue](x) at (10.0,2.5) {$x$};
    	\node[purple](y) at (10.0,0.0) {y/yellow};
    	\node[white](l) at (8.0,1.25) {N\fs N~~~:};
    	\draw[->, thick] (x.south) -- (y.north) node[midway, above, sloped] {:mod};
	\end{scope}
\end{tikzpicture}}
\caption{\raggedright\footnotesize ``yellow''; \SEM{($\svar{1}$~:mod y/yellow)}}
\label{yellow}
\end{subfigure}
\begin{subfigure}[b]{0.22\textwidth}
\resizebox{!}{2cm}{
\begin{tikzpicture}[
white/.style={rectangle, minimum size=7mm},
blue/.style={rectangle, draw=blue!60, fill=blue!5, very thick, minimum size=7mm},
purple/.style={ellipse, draw=purple!60, fill=purple!5, very thick, minimum size=8mm},
purple-dot/.style={rectangle, draw=purple!60, fill=purple!5, very thick, dotted, minimum size=7mm},
]
    	\node[white](a) at (7, 1.15) {(S\bs NP)\bs(S\bs NP)$_2$\fs NP$_1$~~~:};
    	\node[blue](p) at (10.0, 2.5) {$x_2$};
    	\node[blue](y) at (10,0.0) {$x_1$};
    	\draw[->, thick] (p.south) -- (y.north) node[midway, above, sloped] {:location};
\end{tikzpicture}}
\caption{\raggedright\footnotesize ``at''; \SEM{($\svar{2}$ :location $\svar{1}$)}}
\label{at}
\end{subfigure}
\begin{subfigure}[b]{0.13\textwidth}
\resizebox{2cm}{!}{
\begin{tikzpicture}[
blue/.style={rectangle, draw=blue!60, fill=blue!5, very thick, minimum size=7mm},
purple/.style={ellipse, draw=purple!60, fill=purple!5, very thick, minimum size=8mm},
white/.style={rectangle, minimum size=7mm},
]
    \begin{scope}[scale=0.7, transform shape]
	\node[blue](x) at (10.0,1.15) {$x$};
	\node[white](l) at (8.0,1.15) {NP\fs N~~~:};
	\node[white](l2) at (8.0,0.0) {};
	\end{scope}
	
\end{tikzpicture}}
\caption{\raggedright\footnotesize ``the''; \SEM{$\semid$}}
\label{the}
\end{subfigure}
\caption{Linguistic examples as AMR subgraphs: (a) transitive verb, (b) adjective, (c) preposition (in VP adjunct), (d) determiner (identity semantics).}
\end{figure}


\noindent\textbf{Transitive and Intransitive Verbs.}
\Cref{read} shows the semantics for a transitive verb. Since ``read'' has more than one semantic argument, the order of free variables matters: $\svar{1}$, the first free variable, must correspond to NP$_1$, the rightmost syntactic argument in the category.

\noindent\textbf{Adjectives.}
\Cref{yellow} shows the semantics for an adjective. Note that, unlike in the examples above, the root of this subgraph is a free variable, since the root of this subgraph is what will be filled in. Ordinary adverbs have similar semantics.

\noindent\textbf{Prepositional Phrases (Adjunct).}\label{ppadjuncts}
\Cref{at} shows semantics for the locative preposition ``at''.
To derive a prepositional phrase, assume available constituents ``at'': \SEM{($\svar{2}$ :location $\svar{1}$)} and ``the library'': \SEM{(l/library)}, which may be combined by application. 

\noindent\textbf{Null Semantics: Articles, etc.}\label{sec:id}
Some linguistic features, including tense and definite/indefinite articles, are not represented in AMR. For CCG derivations to deal with these elements, there will need to be a semantic representation which 
allows them to be ``syntactic sugar'', affecting the syntactic category but adding nothing to the semantics in the derivation.
We call this the \textbf{identity function}, following \citet{beschke2018graph}, and notate it as \semid. More precisely, if a constituent $a$ has \semid as its semantics, then $a$, when combined with another constituent $b$ via application or composition (either as the function or as the argument), will produce $b$'s semantics for the resulting constituent.

\begin{SCfigure}[.5][b]
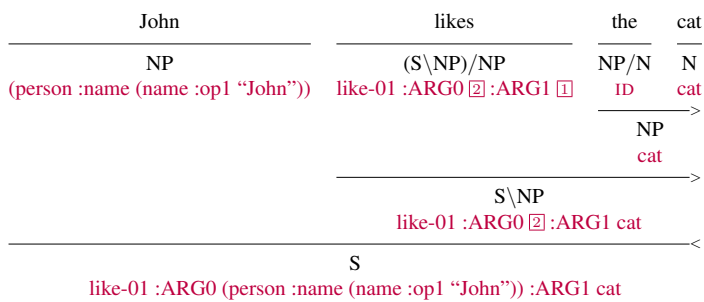
\small
\centering
\resizebox{0.6\columnwidth}{!}{
\begin{tabular}{c c c c}
John & likes & the & cat\\[-1ex]
 \ccgline{1}{} &  \ccgline{1}{} &  \ccgline{1}{} & \ccgline{1}{}\\
 \phrase{1}{NP} &  \phrase{1}{(S\bs NP)\fs NP} & \phrase{1}{NP\fs N}& \phrase{1}{N} \\
 \sem{1}{(person :name (name :op1 ``John''))} &  \sem{1}{like-01 :ARG0 \svar{2} :ARG1 $\svar{1}$} &  \sem{1}{\semid} & \sem{1}{cat} \\[-1ex]
 && \ccgline{2}{$>$}\\
 && \phrase{2}{NP}\\
 && \sem{2}{cat}\\[-1ex]
 &  \ccgline{3}{$>$}\\
 & \phrase{3}{S\bs NP}\\
 & \sem{3}{like-01 :ARG0 $\svar{2}$ :ARG1 cat}\\[-1ex]
 \ccgline{4}{$<$}\\
 \phrase{4}{S}\\
 \sem{4}{like-01 :ARG0 (person :name (name :op1 ``John'')) :ARG1 cat}\\[-4ex]
\end{tabular}
}
\caption{\textbf{application and identity}: ``John likes the cat''}
\label{like-cats}
\end{SCfigure}
\begin{figure}[tb]
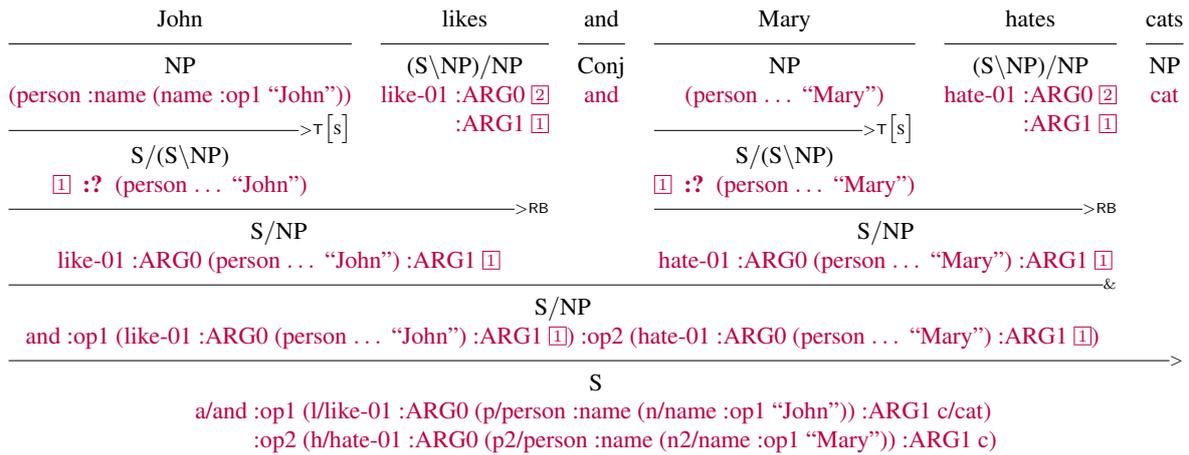
\small
\begin{center}
\resizebox{\columnwidth}{!}{
\begin{tabular}{c c c c c c}
John & likes & and & Mary & hates & cats\\[-1ex]
 \ccgline{1}{} &  \ccgline{1}{} &  \ccgline{1}{} & \ccgline{1}{} & \ccgline{1}{} & \ccgline{1}{}\\
 \phrase{1}{NP} &  \phrase{1}{(S\bs NP)\fs NP} & \phrase{1}{Conj}& \phrase{1}{NP} &  \phrase{1}{(S\bs NP)\fs NP} &  \phrase{1}{NP}\\
 \sem{1}{(person :name (name :op1 ``John''))} &  \sem{1}{\multirow{2}{*}{\begin{tabular}{@{}r@{~}l@{}}like-01 & :ARG0 \svar{2} \\ & :ARG1 $\svar{1}$\end{tabular}}} & \sem{1}{and} & \sem{1}{(person $\dots$ ``Mary'')} & \sem{1}{\multirow{2}{*}{\begin{tabular}{@{}r@{~}l@{}}hate-01 & :ARG0 \svar{2} \\ & :ARG1 $\svar{1}$\end{tabular}}} & \sem{1}{cat} \\
 \ccgline{1}{$>$\combt\big[S\big]}&&& \ccgline{1}{$>$\combt\big[S\big]}\\
 \phrase{1}{S\fs (S\bs NP)}&&&\phrase{1}{S\fs (S\bs NP)}\\
 \sem{1}{\svar{1}~~\qrel~~(person $\dots$ ``John'')}&&&\sem{1}{\svar{1}~~\qrel~~(person $\dots$ ``Mary'')}\\[-1ex]
 \ccgline{2}{$>$\combsb}&&\ccgline{2}{$>$\combsb}\\
 \phrase{2}{S\fs NP}&&\phrase{2}{S\fs NP}\\ 
  \sem{2}{like-01 :ARG0 (person $\dots$ ``John'') :ARG1 \svar{1}}&&\sem{2}{hate-01 :ARG0 (person $\dots$ ``Mary'') :ARG1 \svar{1}}\\[-1ex] 
  \ccgline{5}{$\&$}\\
  \phrase{5}{S\fs NP}\\
  \sem{5}{and :op1 (like-01 :ARG0 (person $\dots$ ``John'') :ARG1 \svar{1}) :op2 (hate-01 :ARG0 (person $\dots$ ``Mary'') :ARG1 \svar{1})}\\[-1ex]
  \ccgline{6}{$>$}\\
  \phrase{6}{S}\\
  \sem{6}{\begin{tabular}{@{}r@{~}l@{}}a/and & :op1 (l/like-01 :ARG0 (p/person :name (n/name :op1 ``John'')) :ARG1 c/cat) \\ & :op2 (h/hate-01 :ARG0 (p2/person :name (n2/name :op1 ``Mary'')) :ARG1 c)\end{tabular}}\\[-4ex]
\end{tabular}
}
\end{center}
\caption{\textbf{complex coordination and type raising}: ``John likes and Mary hates cats''}
\label{coordination}
\end{figure}

\Cref{like-cats} shows the use of application (and identity application) combinators to derive a simple sentence. \Cref{coordination} demonstrates type raising, relation-wise composition, and conjunction as tools to derive a sentence with complex coordination.

\begin{figure}[tb]
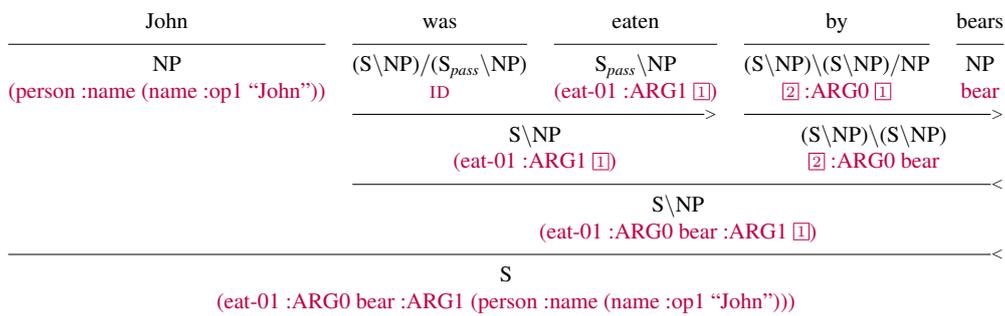
\small
\begin{center}
\resizebox{0.85\columnwidth}{!}{
\begin{tabular}{c c c c c}
John & was & eaten & by & bears\\[-1ex]
 \ccgline{1}{} &  \ccgline{1}{} &  \ccgline{1}{} &  \ccgline{1}{} &  \ccgline{1}{}  \\
 \phrase{1}{NP} &  \phrase{1}{(S\bs NP)\fs (S$_{\textit{pass}}$\bs NP)} &  \phrase{1}{S$_{\textit{pass}}$\bs NP} &  \phrase{1}{(S\bs NP)\bs (S\bs NP)\fs NP} & \phrase{1}{NP} \\
 \sem{1}{(person :name (name :op1 ``John''))} &  \sem{1}{\semid} &  \sem{1}{(eat-01 :ARG1 $\svar{1}$)} &  \sem{1}{$\svar{2}$ :ARG0 $\svar{1}$} &  \sem{1}{bear} \\[-1ex]
 &\ccgline{2}{$>$}&\ccgline{2}{$>$}\\
 & \phrase{2}{S\bs NP}&\phrase{2}{(S\bs NP)\bs (S\bs NP)}\\
 & \sem{2}{(eat-01 :ARG1 $\svar{1}$)} & \sem{2}{$\svar{2}$ :ARG0 bear}\\[-1ex]
 &\ccgline{4}{$<$}\\
 &\phrase{4}{S\bs NP}\\
 &\sem{4}{(eat-01 :ARG0 bear :ARG1 $\svar{1}$)}\\[-1ex]
 \ccgline{5}{$<$}\\
 \phrase{5}{S}\\
 \sem{5}{(eat-01 :ARG0 bear :ARG1 (person :name (name :op1 ``John'')))}\\[-4ex]
\end{tabular}
}
\end{center}
\caption{\textbf{passive}: ``John was eaten by bears''}
\label{passive}
\end{figure}

\begin{figure}[tb]
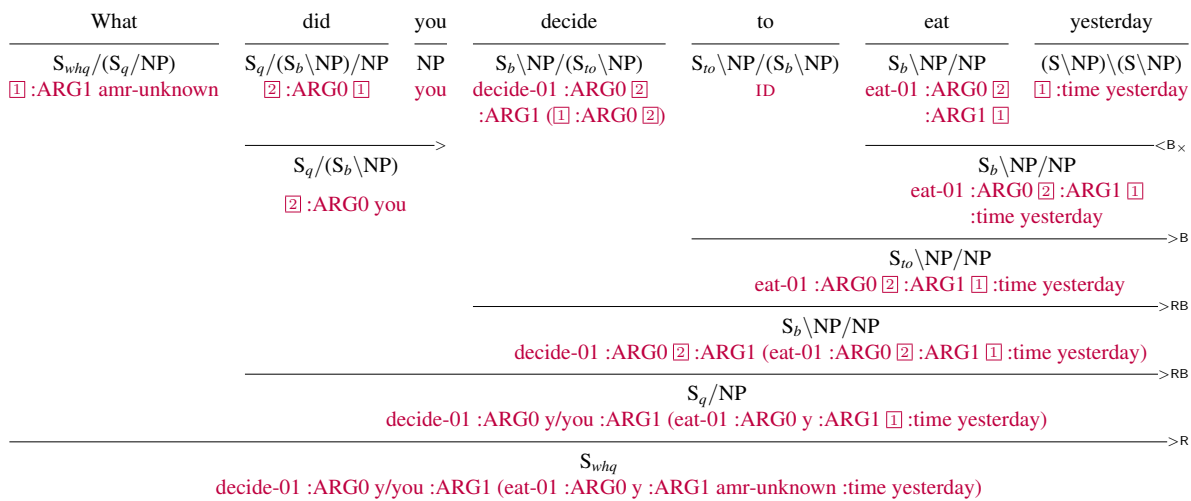
\small
\begin{center}
\resizebox{\columnwidth}{!}{
\begin{tabular}{c c c c c c c}
What & did & you & decide & to & eat & yesterday\\[-1ex]
 \ccgline{1}{} &  \ccgline{1}{} &  \ccgline{1}{} &  \ccgline{1}{} &  \ccgline{1}{} &  \ccgline{1}{} &  \ccgline{1}{} \\
 \phrase{1}{S$_{\textit{whq}}$\fs (S$_{\textit{q}}$\fs NP)} &  \phrase{1}{S$_{\textit{q}}$\fs (S$_{\textit{b}}$\bs NP)\fs NP} &  \phrase{1}{NP} &  \phrase{1}{S$_{\textit{b}}$\bs NP\fs (S$_{\textit{to}}$\bs NP)} &  \phrase{1}{S$_{\textit{to}}$\bs NP\fs (S$_{\textit{b}}$\bs NP)} &  \phrase{1}{S$_{\textit{b}}$\bs NP\fs NP} & \phrase{1}{(S\bs NP)\bs (S\bs NP)} \\
 \sem{1}{$\svar{1}$ :ARG1 amr-unknown} &  \sem{1}{$\svar{2}$ :ARG0 $\svar{1}$} &  \sem{1}{you} & \sem{1}{\multirow{2}{*}{\begin{tabular}{@{}l@{}}decide-01 :ARG0 $\svar{2}$ \\ ~:ARG1 ($\svar{1}$ :ARG0 $\svar{2}$)\end{tabular}}} & \sem{1}{\semid} & \sem{1}{\multirow{2}{*}{\begin{tabular}{@{}r@{~}l@{}}eat-01 & :ARG0 \svar{2} \\ & :ARG1 $\svar{1}$\end{tabular}}} &  \sem{1}{$\svar{1}$ :time yesterday} \\[2.5ex]
 &\ccgline{2}{$>$}& & & \ccgline{2}{$<$\combbx}\\
  &\phrase{2}{S$_{\textit{q}}$\fs (S$_{\textit{b}}$\bs NP)}& & & \phrase{2}{S$_{\textit{b}}$\bs NP\fs NP} \\
  &\sem{2}{$\svar{2}$ :ARG0 you} & & &\sem{2}{\begin{tabular}{@{}r@{~}l@{}}eat-01 & :ARG0 $\svar{2}$ :ARG1 $\svar{1}$ \\ & :time yesterday\end{tabular}}\\[-1ex]
 & & & & \ccgline{3}{$>$\combb}\\
  & & & & \phrase{3}{S$_{\textit{to}}$\bs NP\fs NP} \\
  & & & &\sem{3}{eat-01 :ARG0 $\svar{2}$ :ARG1 $\svar{1}$ :time yesterday}\\[-1ex]
 & & & \ccgline{4}{$>$\combsb}\\
  & & & \phrase{4}{S$_{\textit{b}}$\bs NP\fs NP} \\
  & & &\sem{4}{decide-01 :ARG0 $\svar{2}$ :ARG1 (eat-01 :ARG0 $\svar{2}$ :ARG1 $\svar{1}$ :time yesterday)}\\[-1ex]
  & \ccgline{6}{$>$\combsb}\\
  & \phrase{6}{S$_{\textit{q}}$\fs NP}\\
  & \sem{6}{decide-01 :ARG0 y/you :ARG1 (eat-01 :ARG0 y :ARG1 $\svar{1}$ :time yesterday)}\\[-1ex]
  \ccgline{7}{$>$\combsa}\\
  \phrase{7}{S$_{\textit{whq}}$}\\
  \sem{7}{decide-01 :ARG0 y/you :ARG1 (eat-01 :ARG0 y :ARG1 amr-unknown :time yesterday)}\\[-3ex]
\end{tabular}
}
\end{center}
\caption{\textbf{wh-question and control, relation-wise and non--relation-wise composition}: ``What did you decide to eat yesterday?'' \combbx~stands for \emph{crossing composition}, which has the same semantics as composition.}
\label{wh-control}
\end{figure}

\noindent\textbf{Passives, Control, and Wh-questions.}
\Cref{passive,wh-control} show CCG derivations with AMR semantics for three well-known linguistic phenomena in English: passives, control, and wh-questions. In a passive construction, a semantically core argument may be added by a syntactically optional adjunct phrase
as in \cref{passive}. Note that in this semantic representation, only syntactically required arguments are represented in a predicate's semantics, and so the passive verb \textit{eaten} does not include an \SEM{:ARG0} edge. 

\Cref{wh-control} shows both control and wh-question formation. Control is an important problem for graph semantics as it requires representing the subject (here \textit{you}) as the agent of two predicates (see \cref{sec:sa}). \mbox{Wh-questions} are another complex and difficult phenomenon that is handled by CCG derivation. Additionally, \cref{wh-control} gives examples of both types of composition: relation-wise and non--relation-wise.

\subsection{Inverse Core Roles and Relative Clauses}

AMR provides notation for \emph{inverse roles} that reverse the usual ordering of a relation. 
These are indicated with the \SEM{-of} suffix: \SEM{(a :rel-of b)} is equivalent to \SEM{(b :rel a)}.
This ensures that the graph can be constructed with a single root, and provides a convenient mechanism for expressing derived nominals and relative clauses.
For instance, the noun phrases ``teacher'' and ``a person who teaches'' both receive the AMR \SEM{(person :ARG0-of teach-01)}.
If the subject matter is expressed, that is slotted into the \SEM{:ARG1} of \SEM{teach-01}. This can be handled by treating ``teachers'' as a predicate of sorts, as seen in the derivation below.


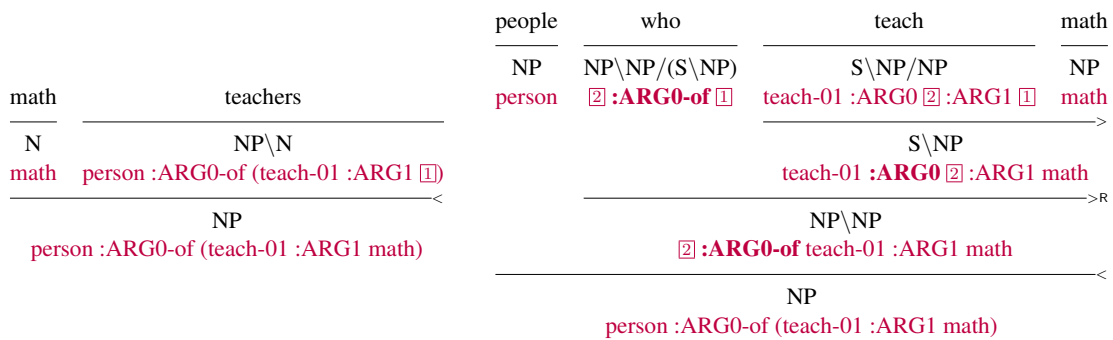
\begin{center}
\resizebox{0.95\columnwidth}{!}{
\begin{tabular}{cc}
    \begin{tabular}{c c}
    math & teachers\\[-1ex]
     \ccgline{1}{} &  \ccgline{1}{} \\
     \phrase{1}{N} &  \phrase{1}{NP\bs N} \\
     \sem{1}{math} &  \sem{1}{person :ARG0-of (teach-01 :ARG1 $\svar{1}$)} \\[-1ex]
     \ccgline{2}{$<$}\\
     \phrase{2}{NP}\\
      \sem{2}{person :ARG0-of (teach-01 :ARG1 math)}\\
    \end{tabular}
&
    \begin{tabular}{c c c c}
    people & who & teach & math\\[-1ex]
     \ccgline{1}{} &  \ccgline{1}{} &  \ccgline{1}{} & \ccgline{1}{}\\
     \phrase{1}{NP} &  \phrase{1}{NP\bs NP\fs (S\bs NP)} & \phrase{1}{S\bs NP\fs NP}& \phrase{1}{NP} \\
     \sem{1}{person} & \sem{1}{\svar{2} \textbf{:ARG0-of} \svar{1}} & \sem{1}{teach-01 :ARG0 \svar{2} :ARG1 \svar{1}} &  \sem{1}{math} \\[-1ex]
     && \ccgline{2}{$>$}\\
     && \phrase{2}{S\bs NP}\\
     && \sem{2}{teach-01 \textbf{:ARG0} \svar{2} :ARG1 math}\\[-1ex]
     &  \ccgline{3}{$>$\combsa}\\
     & \phrase{3}{NP\bs NP}\\
     & \sem{3}{\svar{2} \textbf{:ARG0-of} teach-01 :ARG1 math}\\[-1ex]
     \ccgline{4}{$<$}\\
     \phrase{4}{NP}\\
     \sem{4}{person :ARG0-of (teach-01 :ARG1 math)}\\
    \end{tabular}
\end{tabular}}
\end{center}

Also illustrated is the relative clause paraphrase, ``people who teach math''.
Here, the relativizer ``who'' needs to fill the appropriate role of the verbal predicate with its noun head ``people''. 
An inverse role is produced so that \SEM{person}, rather than \SEM{teach-01}, will be the root of the resulting subgraph.
The relation-wise application combinator must therefore be aware of inverses: it must match the \SEM{:ARG0-of} with the \SEM{:ARG0} edge in the operand and effectively merge the two relations.
Alternatively, the phrase could be parsed by first relation-wise composing ``who'' with ``teach'', which requires similar handling of the inverse role, and then attaching ``math'' by application.

\subsection{Eventive Nouns and PP Complements}\label{challenge}

This section will describe an approach to the semantics of eventive nouns like ``decision'', and in the process will illustrate our treatment of prepositional phrase complements (as opposed to adjuncts: beginning of \cref{ppadjuncts}), which in CCG are traditionally given the category PP.

In English, many eventive nouns can be linked to semantic arguments via prepositional phrases, possessives, and light verb constructions, as shown in \cref{eventive}.
AMR uses a canonical form with a predicate (typically based on a verbal paraphrase), treating \textit{John decided}, \textit{John's decision}, and \textit{John made a/his decision} as semantically equivalent.
Despite some work on integrating event nominals and multiword expressions into CCG \citep{constable-09,honnibal-10,de2015ccg}, we are not aware of any CCG analyses of \textbf{light verb constructions}, which have been studied computationally in other frameworks \citep[e.g.,][]{baldwin-10,bonial-14,ramisch-18}, that gives them semantics equivalent to a content verb paraphrase. 
We offer such an analysis based on three principles:



\begin{enumerate}
\item The event frame is in the semantics of the eventive noun or verb.
\item For any syntactic argument of a noun or verb, the corresponding edge (and free variable) is in the semantics of the noun or verb.
\item Any function word (light verb, \textit{'s}, preposition, or infinitival \textit{to}) that links the eventive noun to its semantic argument has an associated edge (and free variables) in its semantics.
\end{enumerate}

\begin{table}[tb]\small
    \centering
    \begin{tabular}{c|c|c}
         \textbf{light verb construction} & \textbf{possessive form} & \textbf{AMR predicate}\\
         \hline
         make a \textbf{decision} \small{\big[about/on\big]} & my \textbf{decision} \small{\big[about/on\big]} & \SEM{decide-01} \\ 
         pay \textbf{attention} \small{\big[to\big]}& my \textbf{attention} \small{\big[to\big]}& \SEM{attend-02} \\
         make an \textbf{attempt} \small{\big[to\big]}& my \textbf{attempt} \small{\big[to\big]} & \SEM{attempt-01} \\ 
         take a \textbf{nap} & my \textbf{nap} & \SEM{nap-01} \\
         take a \textbf{picture} \small{\big[of\big]}& --- (``my picture'' is not eventive) & \SEM{photograph-01} \emph{(suggested)} \\
    \end{tabular}
    \caption{English eventive nouns shown with a light verb or possessive; words in square brackets mark additional semantic arguments. (In the AMR corpus, ``take pictures'' is actually treated superficially with \SEM{take-01 :ARG1 picture}, but we suggest \SEM{photograph-01} instead.)}
    \label{eventive}
\end{table}
Note that when a verb or noun takes a PP complement, principles 2 and 3 force both the verb or noun and the preposition to hold the same edge in their semantics. This is compatible with relation-wise combinators as described in \cref{sec:sa}. 
The result is a nice analysis where both the eventive noun or verb and its complement preposition signal \textit{patientness}.

With this analysis, the associated light verbs given in \cref{eventive} (``make'', ``pay'', etc.)\ as well as possessive \textit{'s} take the semantics \SEM{\svar{1} :ARG0 \svar{2}}, and associated prepositions take the semantics \SEM{\svar{2} :ARG1 \svar{1}}. In other words, for each eventive noun, either a special light verb or a possessive contributes the agentive semantic relation---and (if present) a special preposition or infinitive \textit{to} may contribute the patient semantic relation---thus allowing derivation of the same AMR regardless of form.

\Cref{decide3} shows the derivation for ``decision'' in its light verb construction form. The preposition ``on'' redundantly represents the \SEM{:ARG1} edge, and is merged with ``decision'' by relation-wise application.\footnote{\label{fn:particle}The category N\fs NP\bs (N\fs PP$_{\textit{on}}$) for ``on'' is suggested by Mark Steedman's analysis of English prepositions as particles (personal communication) and also maintains the Principle of Functional Isomorphism of \cref{sec:iso}.}
The light verb ``made'' specifies the \SEM{:ARG0} edge. 

\begin{figure}[tb]\small
\begin{center}
\resizebox{0.8\columnwidth}{!}{
\begin{tabular}{c c c c c c c}
 John & made & a & decision & on & his & major\\[-1ex]
 \ccgline{1}{} &  \ccgline{1}{} &  \ccgline{1}{} &  \ccgline{1}{} &  \ccgline{1}{} &  \ccgline{1}{} &  \ccgline{1}{}  \\
 \phrase{1}{NP} &\phrase{1}{S\bs NP\fs NP}& \phrase{1}{NP\fs N} &\phrase{1}{N\fs PP$_{\textit{on}}$} &  \phrase{1}{N\fs NP\bs (N\fs PP$_{\textit{on}}$)} &  \phrase{1}{NP\fs N} &  \phrase{1}{N} \\
 \sem{1}{(person :name ~~~~~~~~~~~} &\sem{1}{$\svar{1}$~:ARG0~$\svar{2}$}& \sem{1}{\semid} & \sem{1}{(decide-01 :ARG1 $\svar{1}$)} &  \sem{1}{$\svar{2}$~:ARG1~$\svar{1}$} &  \sem{1}{$\svar{1}$ :poss he} &  \sem{1}{major} \\
  \sem{1}{(name :op1 ``John''))}&&& &&&\\[-1ex]
 &&&\ccgline{2}{$<$\combsa}  &\ccgline{2}{$>$}\\
  &&&\phrase{2}{N\fs NP}&\phrase{2}{NP}\\
    &&&\sem{2}{(decide-01 :ARG1 $\svar{1}$)} &  \sem{2}{(major :poss he)} \\[-1ex]
    &&&\ccgline{4}{$>$}\\
    &&&\phrase{4}{N}\\
    &&&\sem{4}{(decide-01 :ARG1~(major :poss he))}\\[-1ex]
 &&\ccgline{5}{$>$}\\
 & &\phrase{5}{NP}\\
 & &  \sem{5}{(decide-01 :ARG1~(major :poss he))} \\[-1ex]
 & \ccgline{6}{$>$}\\
  & \phrase{6}{S\bs NP}\\
   & \sem{6}{(decide-01 :ARG0 $\svar{2}$ :ARG1~(major :poss he))}\\[-1ex]
\ccgline{7}{$<$}\\
\phrase{7}{S}\\
\sem{7}{(decide-01 :ARG0 (person :name (name :op1 ``John'')) :ARG1~(major :poss he))} 
\\[-4ex]
\end{tabular}
}
\end{center}
\caption{\textbf{light verb construction}: ``John made a decision on his major''}
\label{decide3}
\end{figure}



\section{Discussion}\label{sec:discussion}

Unlike many semantic formalisms, AMR does not specify a `compositional story': annotations do not include any sort of syntactic derivation, or even gold alignments between semantic units and words in the sentence. 
This presents a challenge for AMR parsing, which in practice relies on various forms of automatic or latent alignments \citep[see][]{szubert-18}.
Above, we have presented an analysis that lays the foundation for a linguistically principled treatment of CCG-to-AMR parsing that meets a variety of challenges in the syntax-semantics interface, and does so in a transparent way so that parsing errors can be diagnosed.
We believe the approach is reasonably intuitive, flowing naturally from CCG syntax, AMR semantics, and the notion of free variables in subgraphs, without the additional need for complicated lambda calculus notation or a highly general graph grammar formalism.

To realize this vision in practice, an approach is needed to build a CCG parser enriched with graph semantics for deriving AMRs. 
We anticipate that existing CCG parsing frameworks can be adapted---for example, by developing an alignment algorithm to induce the semantics for lexical entries from the AMR corpus, and running an off-the-shelf parser like EasySRL \citep{lewis-15} at training and test time for the syntactic side of the derivation. 
This approach would take advantage of the fact that our analysis assumes the ordinary CCG syntax for obtaining the compositional structure of the derivation. 
The only additional steps would be a)~disambiguating the semantics of lexical entries in the derivation, and b)~applying the semantics of the combinators as specified in \cref{all-combinators}. 
For each use of application or composition, the semantic parser would check whether the conditions for relation-wise combination hold, and otherwise apply the ordinary version of the combinator.\footnote{\label{fn:underspecified}We have considered an alternative analysis where underspecified \qrel edges would be used not only for type raising, but for all case-marked pronouns, prepositions marking syntactic arguments, and other constructions where a word's syntactic category involves an argument to a separate predicate. Thus, only the predicate would be allowed to specify semantic roles for its syntactic arguments. Relation-wise combinators would then require that the shared edge would be underspecified in the function constituent. The rationale would be that this avoids redundant specification of core roles like \SEM{:ARG0} and \SEM{:ARG1} in the lexical entries---e.g.~in \cref{wh-control}, the \SEM{:ARG1} for ``What'', the \SEM{:ARG0} for ``did'', and the second \SEM{:ARG0} for ``decide'' would all be replaced with \qrel. After all, constructions like wh-questions, control, and case target syntactic relations (subject\slash object), which are merely \emph{correlated} with semantic roles. And as pointed out by a reviewer, under the current approach, a wrong choice of semantic role for a cased pronoun's semantics could result in the use of a regular combinator rather than a relation-wise combinator, leaving a free variable in the predicate unsatisfied and essentially breaking the syntax-semantics isomorphism. An argument in favor of the current policy is that prepositions can contain information about roles to a certain extent, and redundant specification of semantic roles may actually be helpful when confronted with a noisy parser and lexicon. We leave this open as an empirical question for parsing research.}

Because AMRs are annotated by humans for raw sentences, rather than on top of a syntactic parse, we cannot expect a parser to elegantly handle the full construction of all AMRs according to compositional rules.
Several components of AMR parsing are not part of CCG parsing and will have to be performed as postprocessing steps. These components include named entity recognition, time expression parsing, coreference resolution, and wikification, all of which need to be performed after (or before) CCG parsing. Additionally, there is a risk that a CCG lexicon may `overgenerate', producing invalid parses, and additional checking---either in the combinators, or as postprocessing or reranking---may be warranted.


We are aware of certain phenomena where the approach described above would be unable to fully match the conventions of AMR in the CCG-derived semantics. 
The analysis presented for \textbf{coordination} (with the conjunction combinator: see \cref{coordination}) 
would address only one of the ways it can be expressed in AMR, with a concept like \SEM{and} or \SEM{or} and operands. 
In other cases, coordinated modifiers are treated as sister relations in the AMR, with no explicit concept for the conjunction. 
Even when the conjunction is explicit in the AMR, it may be placed at a different level in the gold and CCG-derived AMRs: e.g., when two purpose adjuncts are coordinated, the derivation will result in semantic conjunction over the predicate rather than a conjunction under the \SEM{:purpose} relation. 
In sentences where a semantic predicate is \textbf{duplicated} in the AMR with different participants, e.g.~due to right node raising, a copy mechanism would be needed to avoid spurious reentrancy. 
The treatment of \textbf{modal auxiliaries} as above the main event predicate in the AMR will be problematic for the CCG derivation when there is a preposed adjunct (as in ``\emph{Tomorrow}, John may eat rice'') because the modifier will semantically attach under the root of the semantics of the rest of the clause (\SEM{possible-01} from ``may'') rather than the main event predicate \SEM{eat-01}. 
Full derivations for these problem cases, as well as examples of purpose clauses, raising, and subject and object control, are given in  \cref{appendix}.
We will explore whether such limitations can be addressed via postprocessing of the parse, or whether additional expressive power in the combinators is necessary.

Finally, as pointed out by \citet{bender-15}, AMR annotations sometimes go beyond the compositional `sentence meaning' and incorporate elements of `speaker meaning', though an empirical study of AMR data found the rate of noncompositional structures to be relatively low \citep{szubert-18}. 
\Citet{beschke2018graph} give interesting examples of AMR fragments that would be difficult to derive compositionally, e.g., ``settled on Indianapolis for its board meeting'', where the AMR attaches Indianapolis as the location of the meeting and the meeting as the thing that was settled on (reflecting the inference \textit{settle on \textsc{location} for \textsc{activity}} $\Rightarrow$ \textit{settle on} [\textsc{activity} \textit{at} \textsc{location}]).



\section{Conclusion}\label{sec:conclusion}
We have given the linguistic motivation for a particular method of deriving AMR semantic graphs using CCG.
Our specification of AMR subgraphs and CCG combinators ensures a tight correspondence between syntax and semantics, which we have illustrated for a variety of linguistic constructions (including light verb construction semantics, which to the best of our knowledge has not previously been explored for CCG).
Future empirical work can make use of this framework to induce CCG lexicons for AMR parsing.


\nonanonversion{\section*{Acknowledgments}
We want to thank Paul Portner, Adam Lopez, members of the NERT lab at Georgetown, and anonymous reviewers for their helpful feedback on this research, as well as Matthew Honnibal, Siva Reddy, and Mark Steedman for early discussions about light verbs in CCG.
}

{\bibliography{mybibfile}
\bibliographystyle{chicago}}

\clearpage

\addtolength{\dbltextfloatsep}{.5cm} 
\addtolength{\intextsep}{.5cm}

\appendix

\section{Additional Derivations}\label{appendix}

Below are full derivations illustrating raising, subject control, object control, an object control wh-question, a modal auxiliary with preposed VP adjunct, a purpose clause, coordinated purpose clauses, and right node raising with a shared main verb.

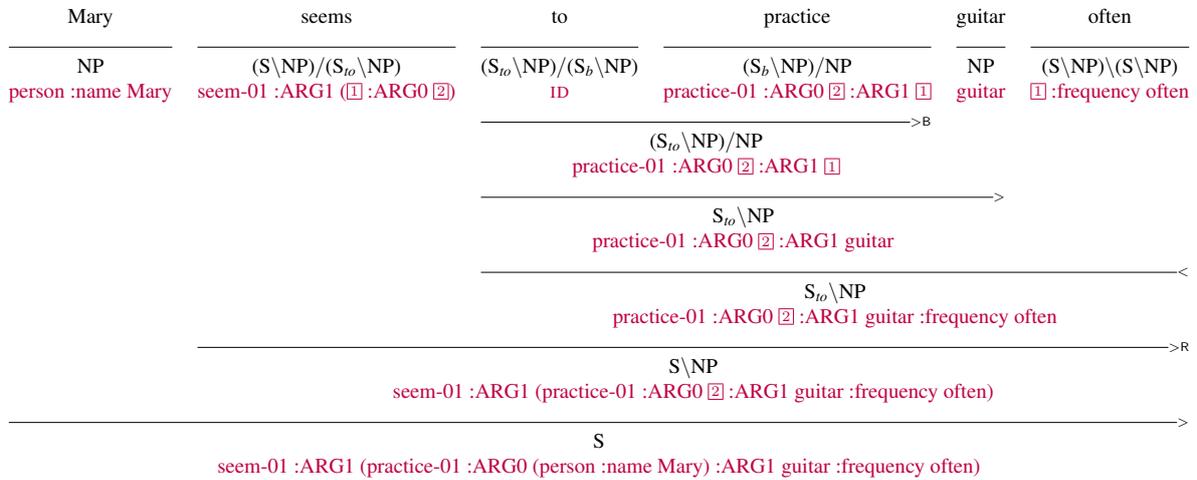
\begin{figure}[htb]\small\centering
\resizebox{\columnwidth}{!}{
\begin{tabular}{c c c c c c}
Mary & seems & to & practice & guitar & often\\
 \ccgline{1}{} &  \ccgline{1}{} &  \ccgline{1}{} &  \ccgline{1}{} &  \ccgline{1}{} & \ccgline{1}{} \\
 \phrase{1}{NP} & \phrase{1}{(S\bs NP)\fs (S$_{\textit{to}}$\bs NP)} & \phrase{1}{(S$_{\textit{to}}$\bs NP)\fs (S$_{\textit{b}}$\bs NP)} & \phrase{1}{(S$_{\textit{b}}$\bs NP)\fs NP} & \phrase{1}{NP} & \phrase{1}{(S\bs NP)\bs (S\bs NP)}\\
 \sem{1}{person :name Mary} & \sem{1}{seem-01 :ARG1 (\svar{1} :ARG0 \svar{2})} & \sem{1}{\semid} & \sem{1}{practice-01 :ARG0 \svar{2} :ARG1 $\svar{1}$} & \sem{1}{guitar} & \sem{1}{\svar{1} :frequency often}\\
 &&\ccgline{2}{$>$\combb}&\\
 &&\phrase{2}{(S$_{\textit{to}}$\bs NP)\fs NP}&\\
 &&\sem{2}{practice-01 :ARG0 \svar{2} :ARG1 $\svar{1}$}&\\ 
 &&\ccgline{3}{$>$}\\
 &&\phrase{3}{S$_{\textit{to}}$\bs NP}\\
 &&\sem{3}{practice-01 :ARG0 \svar{2} :ARG1 guitar}\\ 
  &&\ccgline{4}{$<$}\\
 &&\phrase{4}{S$_{\textit{to}}$\bs NP}\\
 &&\sem{4}{practice-01 :ARG0 \svar{2} :ARG1 guitar :frequency often}\\ 
 &\ccgline{5}{$>$\combsa}\\
 &\phrase{5}{S\bs NP}\\
 &\sem{5}{seem-01 :ARG1 (practice-01 :ARG0 \svar{2} :ARG1 guitar :frequency often)}\\ 
  \ccgline{6}{$>$}\\
 \phrase{6}{S}\\
 \sem{6}{seem-01 :ARG1 (practice-01 :ARG0 (person :name Mary) :ARG1 guitar :frequency often)}\\ 
\end{tabular}
}
\caption{Raising}
\label{raising}
\end{figure}

\begin{figure}[htb]
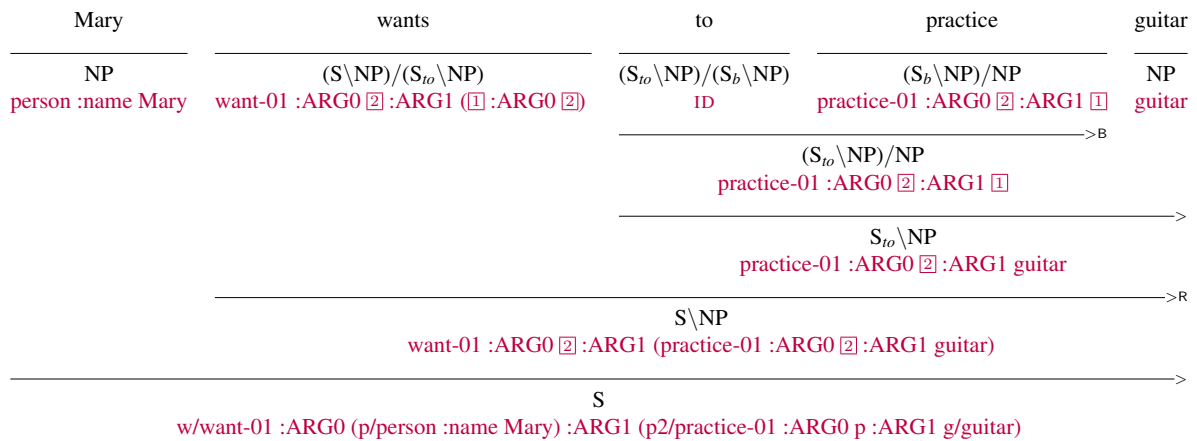
\small\centering
\resizebox{\columnwidth}{!}{
\begin{tabular}{c c c c c}
Mary & wants & to & practice & guitar\\
 \ccgline{1}{} &  \ccgline{1}{} &  \ccgline{1}{} &  \ccgline{1}{} &  \ccgline{1}{} \\
 \phrase{1}{NP} & \phrase{1}{(S\bs NP)\fs (S$_{\textit{to}}$\bs NP)} & \phrase{1}{(S$_{\textit{to}}$\bs NP)\fs (S$_{\textit{b}}$\bs NP)} & \phrase{1}{(S$_{\textit{b}}$\bs NP)\fs NP} & \phrase{1}{NP}\\
 \sem{1}{person :name Mary} & \sem{1}{want-01 :ARG0 \svar{2} :ARG1 ($\svar{1}$ :ARG0 $\svar{2}$) } & \sem{1}{\semid} & \sem{1}{practice-01 :ARG0 \svar{2} :ARG1 $\svar{1}$} & \sem{1}{guitar} \\
  &&\ccgline{2}{$>$\combb}\\
  &&\phrase{2}{(S$_{\textit{to}}$\bs NP)\fs NP}\\
 &&\sem{2}{practice-01 :ARG0 \svar{2} :ARG1 $\svar{1}$}\\ 
 &&\ccgline{3}{$>$}\\
 &&\phrase{3}{S$_{\textit{to}}$\bs NP}\\
 &&\sem{3}{practice-01 :ARG0 \svar{2} :ARG1 guitar}\\ 
 &\ccgline{4}{$>$\combsa}\\
 &\phrase{4}{S\bs NP}\\
 &\sem{4}{want-01 :ARG0 \svar{2} :ARG1 (practice-01 :ARG0 \svar{2} :ARG1 guitar)}\\ 
  \ccgline{5}{$>$}\\
 \phrase{5}{S}\\
 \sem{5}{w/want-01 :ARG0 (p/person :name Mary) :ARG1 (p2/practice-01 :ARG0 p :ARG1 g/guitar)}\\ 
\end{tabular}
}
\caption{Subject control}
\label{subject-control}
\end{figure}

\begin{figure}[htb]
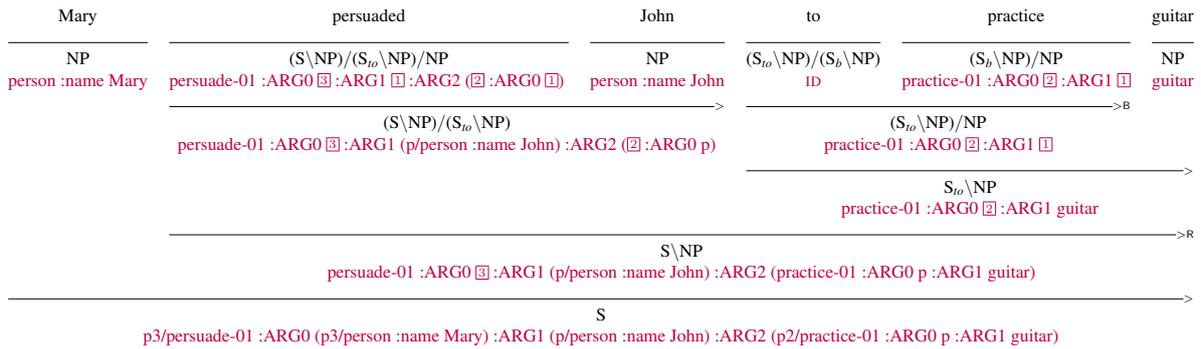
\small\centering
\resizebox{\columnwidth}{!}{
\begin{tabular}{c c c c c c}
Mary & persuaded & John & to & practice & guitar\\
 \ccgline{1}{} &  \ccgline{1}{} &  \ccgline{1}{} &  \ccgline{1}{} &  \ccgline{1}{} &  \ccgline{1}{} \\
 \phrase{1}{NP} & \phrase{1}{(S\bs NP)\fs (S$_{\textit{to}}$\bs NP)\fs NP} & \phrase{1}{NP} & \phrase{1}{(S$_{\textit{to}}$\bs NP)\fs (S$_{\textit{b}}$\bs NP)} & \phrase{1}{(S$_{\textit{b}}$\bs NP)\fs NP} & \phrase{1}{NP} \\
 \sem{1}{person :name Mary} & \sem{1}{persuade-01 :ARG0 \svar{3} :ARG1 \svar{1} :ARG2 ($\svar{2}$ :ARG0 $\svar{1}$) } & \sem{1}{person :name John} & \sem{1}{\semid} & \sem{1}{practice-01 :ARG0 \svar{2} :ARG1 $\svar{1}$} & \sem{1}{guitar}\\
 &\ccgline{2}{$>$}&\ccgline{2}{$>$\combb}\\
 &\phrase{2}{(S\bs NP)\fs (S$_{\textit{to}}$\bs NP)}&\phrase{2}{(S$_{\textit{to}}$\bs NP)\fs NP}\\
 &\sem{2}{persuade-01 :ARG0 \svar{3} :ARG1 (p/person :name John) :ARG2 ($\svar{2}$ :ARG0 p)}&\sem{2}{practice-01 :ARG0 \svar{2} :ARG1 $\svar{1}$}\\ 
 &&&\ccgline{3}{$>$}\\
 &&&\phrase{3}{S$_{\textit{to}}$\bs NP}\\
 &&&\sem{3}{practice-01 :ARG0 \svar{2} :ARG1 guitar}\\ 
 &\ccgline{5}{$>$\combsa}\\
 &\phrase{5}{S\bs NP}\\
 &\sem{5}{persuade-01 :ARG0 \svar{3} :ARG1 (p/person :name John) :ARG2 (practice-01 :ARG0 p :ARG1 guitar)}\\ 
  \ccgline{6}{$>$}\\
 \phrase{6}{S}\\
 \sem{6}{p3/persuade-01 :ARG0 (p3/person :name Mary) :ARG1 (p/person :name John) :ARG2 (p2/practice-01 :ARG0 p :ARG1 guitar)}\\ 
\end{tabular}
}
\caption{Object control. Note that the PropBank predicate \SEM{persuade-01} specifies \SEM{:ARG0} for the persuader, \SEM{:ARG1} for the persuadee, and \SEM{:ARG2} for the impelled action.}
\label{object-control}
\end{figure}

\begin{figure}[htb]
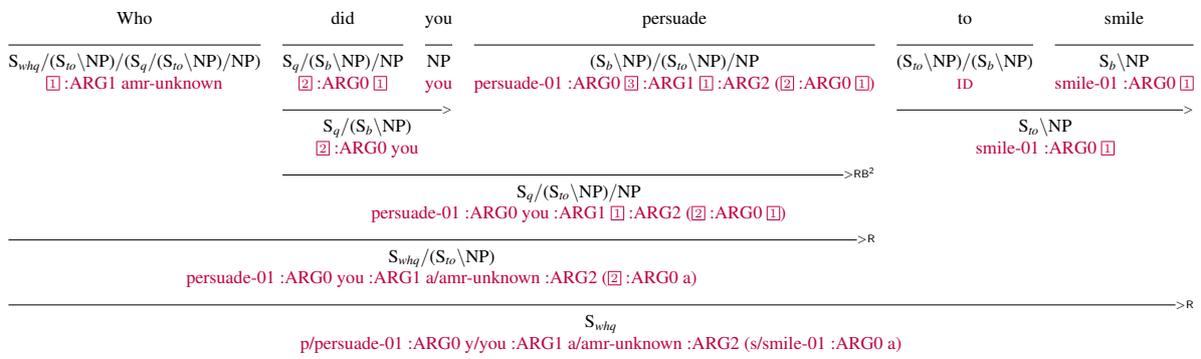
\small
\begin{center}
\resizebox{\columnwidth}{!}{
\begin{tabular}{c c c c c c}
Who & did & you & persuade & to & smile\\
 \ccgline{1}{} &  \ccgline{1}{} &  \ccgline{1}{} &  \ccgline{1}{} &  \ccgline{1}{} & \ccgline{1}{} \\
\phrase{1}{S$_{\textit{whq}}$\fs(S$_{\textit{to}}$\bs NP)\fs (S$_{\textit{q}}$\fs(S$_{\textit{to}}$\bs NP)\fs NP)} &  \phrase{1}{S$_{\textit{q}}$\fs (S$_{\textit{b}}$\bs NP)\fs NP}  & \phrase{1}{NP} & \phrase{1}{(S$_{\textit{b}}$\bs NP)\fs(S$_{\textit{to}}$\bs NP)\fs NP} & \phrase{1}{(S$_{\textit{to}}$\bs NP)\fs (S$_{\textit{b}}$\bs NP)} & \phrase{1}{S$_{\textit{b}}$\bs NP}\\
 \sem{1}{\svar{1}~:ARG1 amr-unknown} & \sem{1}{
 \svar{2} :ARG0 \svar{1}} & \sem{1}{you} & \sem{1}{persuade-01 :ARG0 \svar{3} :ARG1 \svar{1} :ARG2 ($\svar{2}$ :ARG0 $\svar{1}$)} & \sem{1}{\semid} & \sem{1}{smile-01 :ARG0 \svar{1}}\\
 & \ccgline{2}{$>$} && \ccgline{2}{$>$}\\
 & \phrase{2}{S$_{\textit{q}}$\fs (S$_{\textit{b}}$\bs NP)} && \phrase{2}{S$_{\textit{to}}$\bs NP}\\
 & \sem{2}{\svar{2} :ARG0 you} && \sem{2}{smile-01 :ARG0 \svar{1}}\\
 & \ccgline{3}{$>$\combsbb} \\
 & \phrase{3}{S$_{\textit{q}}$\fs(S$_{\textit{to}}$\bs NP)\fs NP} \\
 & \sem{3}{persuade-01 :ARG0 you :ARG1 \svar{1} :ARG2 ($\svar{2}$ :ARG0 $\svar{1}$)} \\
\ccgline{4}{$>$\combsa}\\
\phrase{4}{S$_{\textit{whq}}$\fs(S$_{\textit{to}}$\bs NP)}\\
\sem{4}{persuade-01 :ARG0 you :ARG1 a/amr-unknown :ARG2 ($\svar{2}$ :ARG0 a)}\\
   \ccgline{6}{$>$\combsa}\\
   \phrase{6}{S$_{\textit{whq}}$}\\
   \sem{6}{p/persuade-01 :ARG0 y/you :ARG1 a/amr-unknown :ARG2 (s/smile-01 :ARG0 a)}

 \end{tabular}
}
\end{center}
\caption{Object control wh-question: ``Who did you persuade to smile?'' (example suggested by a reviewer)}
\label{object-control2}
\end{figure}

\begin{figure}[htb]
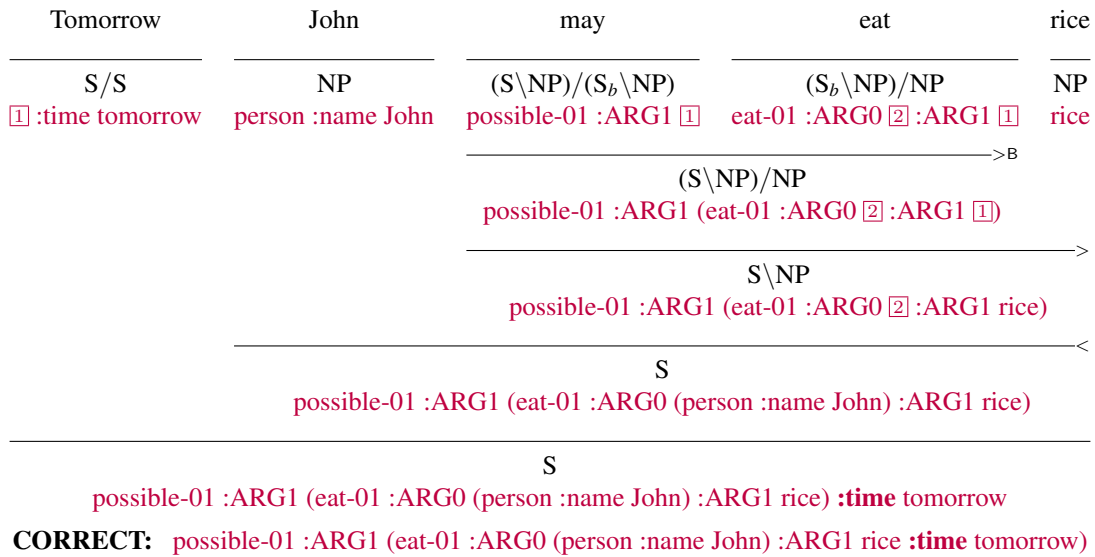
\small
\begin{center}
\begin{tabular}{c c c c c}
Tomorrow & John & may & eat & rice\\
 \ccgline{1}{} &  \ccgline{1}{} &  \ccgline{1}{} &  \ccgline{1}{} &  \ccgline{1}{} \\
\phrase{1}{S\fs S} & \phrase{1}{NP}& \phrase{1}{(S\bs NP)\fs (S$_{\textit{b}}$\bs NP)}& \phrase{1}{(S$_{\textit{b}}$\bs NP)\fs NP}& \phrase{1}{NP} \\ 
\sem{1}{\svar{1} :time tomorrow} & \sem{1}{person :name John} & \sem{1}{possible-01 :ARG1 \svar{1}} & \sem{1}{eat-01 :ARG0 \svar{2} :ARG1 \svar{1}} & \sem{1}{rice}\\
&& \ccgline{2}{$>$\combb}\\
 && \phrase{2}{(S\bs NP)\fs NP}\\
 && \sem{2}{possible-01 :ARG1 (eat-01 :ARG0 \svar{2} :ARG1 \svar{1})}\\
&& \ccgline{3}{$>$}\\
 && \phrase{3}{S\bs NP}\\
 && \sem{3}{possible-01 :ARG1 (eat-01 :ARG0 \svar{2} :ARG1 rice)}\\
& \ccgline{4}{$<$}\\
 & \phrase{4}{S}\\
 & \sem{4}{possible-01 :ARG1 (eat-01 :ARG0 (person :name John) :ARG1 rice)}\\
\ccgline{5}{}\\
\phrase{5}{S}\\
\sem{5}{possible-01 :ARG1 (eat-01 :ARG0 (person :name John) :ARG1 rice) \textbf{:time} tomorrow}\\[1ex]
\sem{5}{\textbf{\color{black}CORRECT:}~~~possible-01 :ARG1 (eat-01 :ARG0 (person :name John) :ARG1 rice \textbf{:time} tomorrow)}
\end{tabular}
\end{center}
\caption{Modal auxiliary with preposed adjunct: ``Tomorrow, John may eat rice''. In the derived AMR, the temporal modifier is placed incorrectly under the modal predicate rather than the main event predicate.}
\label{modal-adjunct}
\end{figure}

\begin{figure}[htb]
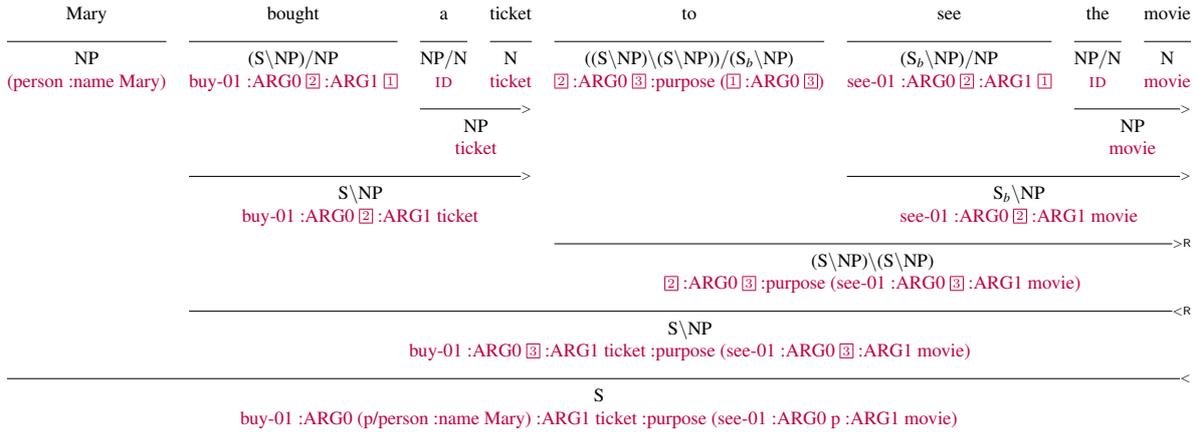
\small
\begin{center}
\resizebox{\columnwidth}{!}{
\begin{tabular}{c c c c c c c c}
Mary & bought & a & ticket & to & see & the & movie\\ 
 \ccgline{1}{} &  \ccgline{1}{} &  \ccgline{1}{} & \ccgline{1}{} &  \ccgline{1}{} &  \ccgline{1}{} & \ccgline{1}{} & \ccgline{1}{}\\
 \phrase{1}{NP} &  \phrase{1}{(S\bs NP)\fs NP} & \phrase{1}{NP\fs N}& \phrase{1}{N} & \phrase{1}{((S\bs NP)\bs (S\bs NP))\fs (S$_{\textit{b}}$\bs NP)} & \phrase{1}{(S$_{\textit{b}}$\bs NP)\fs NP} & \phrase{1}{NP\fs N}& \phrase{1}{N} \\
 \sem{1}{(person :name Mary)} &  \sem{1}{buy-01 :ARG0 \svar{2} :ARG1 $\svar{1}$} &  \sem{1}{\semid} &  \sem{1}{ticket} & \sem{1}{$\svar{2}$ :ARG0 $\svar{3}$ :purpose ($\svar{1}$ :ARG0 $\svar{3}$)} &  \sem{1}{see-01 :ARG0 \svar{2} :ARG1 $\svar{1}$} & \sem{1}{\semid} & \sem{1}{movie}\\
 &&\ccgline{2}{$>$}&&&\ccgline{2}{$>$}\\
 &&\phrase{2}{NP}&&&\phrase{2}{NP}\\
 &&\sem{2}{ticket}&&&\sem{2}{movie}\\ 
 &\ccgline{3}{$>$}&&\ccgline{3}{$>$}\\
  &\phrase{3}{S\bs NP}&&\phrase{3}{S$_{\textit{b}}$\bs NP}\\
 &\sem{3}{buy-01 :ARG0 \svar{2} :ARG1 ticket}&&\sem{3}{see-01 :ARG0 \svar{2} :ARG1 movie}\\
  &&&&\ccgline{4}{$>$\combsa}\\
  &&&&\phrase{4}{(S\bs NP)\bs (S\bs NP)}\\
  &&&&\sem{4}{$\svar{2}$ :ARG0 $\svar{3}$ :purpose (see-01 :ARG0 \svar{3} :ARG1 movie)}\\
 &\ccgline{7}{$<$\combsa}\\
  &\phrase{7}{S\bs NP}\\
  &\sem{7}{buy-01 :ARG0 \svar{3} :ARG1 ticket :purpose (see-01 :ARG0 \svar{3} :ARG1 movie)}\\
  \ccgline{8}{$<$}\\
  \phrase{8}{S}\\
  \sem{8}{buy-01 :ARG0 (p/person :name Mary) :ARG1 ticket :purpose (see-01 :ARG0 p :ARG1 movie)}\\

\end{tabular}
}
\end{center}
\caption{\emph{to}-purpose}
\label{to-purpose}
\end{figure}

\begin{figure}[tbh]
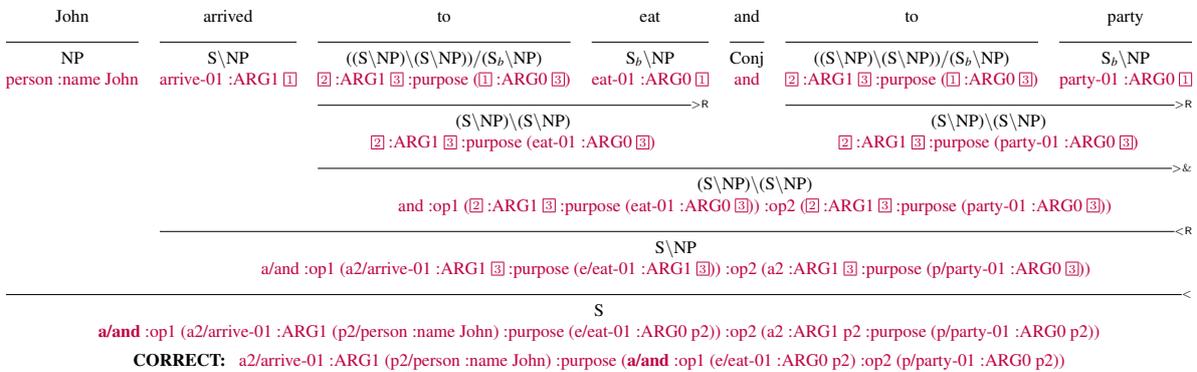
\small
\begin{center}
\resizebox{\columnwidth}{!}{
\begin{tabular}{c c c c c c c}
John & arrived & to & eat & and & to & party\\
 \ccgline{1}{} &  \ccgline{1}{} &  \ccgline{1}{} &  \ccgline{1}{} &  \ccgline{1}{} & \ccgline{1}{} & \ccgline{1}{} \\
\phrase{1}{NP} &  \phrase{1}{S\bs NP}  & \phrase{1}{((S\bs NP)\bs (S\bs NP))\fs (S$_{\textit{b}}$\bs NP)} & \phrase{1}{S$_{\textit{b}}$\bs NP}& \phrase{1}{Conj} & \phrase{1}{((S\bs NP)\bs (S\bs NP))\fs (S$_{\textit{b}}$\bs NP)} & \phrase{1}{S$_{\textit{b}}$\bs NP}\\
\sem{1}{person :name John} & \sem{1}{arrive-01 :ARG1 \svar{1}} & \sem{1}{$\svar{2}$ :ARG1 $\svar{3}$ :purpose ($\svar{1}$ :ARG0 $\svar{3}$)} & \sem{1}{eat-01 :ARG0 \svar{1}} & \sem{1}{and} & \sem{1}{$\svar{2}$ :ARG1 $\svar{3}$ :purpose ($\svar{1}$ :ARG0 $\svar{3}$)} & \sem{1}{party-01 :ARG0 \svar{1}}\\
 &&\ccgline{2}{$>$\combsa}&&\ccgline{2}{$>$\combsa}\\
 &&\phrase{2}{(S\bs NP)\bs (S\bs NP)}&&\phrase{2}{(S\bs NP)\bs (S\bs NP)}\\
 &&\sem{2}{$\svar{2}$ :ARG1 $\svar{3}$ :purpose (eat-01 :ARG0 \svar{3})}&&\sem{2}{$\svar{2}$ :ARG1 $\svar{3}$ :purpose (party-01 :ARG0 \svar{3})}\\
  &&\ccgline{5}{$>$\combconj}\\
  &&\phrase{5}{(S\bs NP)\bs (S\bs NP)}\\
 &&\sem{5}{and :op1 ($\svar{2}$ :ARG1 $\svar{3}$ :purpose (eat-01 :ARG0 \svar{3})) :op2 ($\svar{2}$ :ARG1 $\svar{3}$ :purpose (party-01 :ARG0 \svar{3}))}\\
  &\ccgline{6}{$<$\combsa}\\
  &\phrase{6}{S\bs NP}\\
 &\sem{6}{a/and :op1 (a2/arrive-01 :ARG1 $\svar{3}$ :purpose (e/eat-01 :ARG1 \svar{3})) :op2 (a2 :ARG1 $\svar{3}$ :purpose (p/party-01 :ARG0 \svar{3}))}\\
  \ccgline{7}{$<$}\\
  \phrase{7}{S}\\
 \sem{7}{\textbf{a/and} :op1 (a2/arrive-01 :ARG1 (p2/person :name John) :purpose (e/eat-01 :ARG0 p2)) :op2 (a2 :ARG1 p2 :purpose (p/party-01 :ARG0 p2))}\\[1ex]
 \sem{7}{\textbf{\color{black}CORRECT:}~~~a2/arrive-01 :ARG1 (p2/person :name John) :purpose (\textbf{a/and} :op1 (e/eat-01 :ARG0 p2) :op2 (p/party-01 :ARG0 p2))}
 
\end{tabular}
}
\end{center}
\caption{Coordinated purpose clauses: ``John arrived to eat and to party''. Note that the PropBank predicate \SEM{arrive-01} has no \SEM{:ARG0}; its subject is \SEM{:ARG1}. The lexical semantics for infinitive purpose \emph{to} is chosen accordingly. 
However, the placement in the derived AMR of the semantic conjunction \SEM{and} is incorrect.}
\label{coordinated-purpose}
\end{figure}

\begin{figure}[tbh]
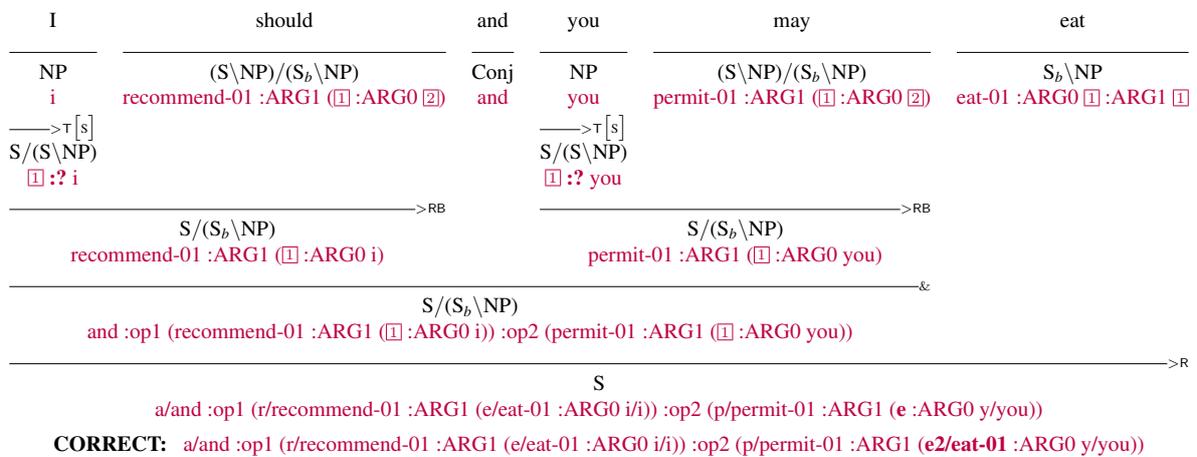
\small
\begin{center}
\resizebox{\columnwidth}{!}{
\begin{tabular}{c c c c c c}
 I & should & and & you & may & eat\\
 \ccgline{1}{} &  \ccgline{1}{} &  \ccgline{1}{} &  \ccgline{1}{} &  \ccgline{1}{} & \ccgline{1}{}  \\
\phrase{1}{NP} &  \phrase{1}{(S\bs NP)\fs(S$_{\textit{b}}$\bs NP)}  & \phrase{1}{Conj} & \phrase{1}{NP}& \phrase{1}{(S\bs NP)\fs(S$_{\textit{b}}$\bs NP)} & \phrase{1}{S$_{\textit{b}}$\bs NP}\\
\sem{1}{i} & \sem{1}{recommend-01 :ARG1 (\svar{1} :ARG0 \svar{2})} & \sem{1}{and} & \sem{1}{you} & \sem{1}{permit-01 :ARG1  (\svar{1} :ARG0 \svar{2})} & \sem{1}{eat-01 :ARG0 \svar{1} :ARG1 \svar{1}}\\
 \ccgline{1}{$>$\combt\big[S\big]}&&&\ccgline{1}{$>$\combt\big[S\big]}\\
 \phrase{1}{S\fs(S\bs NP)}&&&\phrase{1}{S\fs(S\bs NP)}\\
 \sem{1}{\svar{1} \qrel~i}&&&\sem{1}{\svar{1} \qrel~you}\\
 \ccgline{2}{$>$\combsb}&&\ccgline{2}{$>$\combsb}\\
 \phrase{2}{S\fs(S$_{\textit{b}}$\bs NP)}&&\phrase{2}{S\fs(S$_{\textit{b}}$\bs NP)}\\
 \sem{2}{recommend-01 :ARG1 (\svar{1} :ARG0 i)}&&\sem{2}{permit-01 :ARG1 (\svar{1} :ARG0 you)}\\
\ccgline{5}{\combconj}\\
\phrase{5}{S\fs(S$_{\textit{b}}$\bs NP)}\\
\sem{5}{and :op1 (recommend-01 :ARG1 (\svar{1} :ARG0 i)) :op2 (permit-01 :ARG1 (\svar{1} :ARG0 you))}\\
\ccgline{6}{$>$\combsa}\\
\phrase{6}{S}\\
\sem{6}{a/and :op1 (r/recommend-01 :ARG1 (e/eat-01 :ARG0 i/i)) :op2 (p/permit-01 :ARG1 (\textbf{e} :ARG0 y/you))}\\[1ex]
\sem{6}{\textbf{\color{black}CORRECT:}~~~a/and :op1 (r/recommend-01 :ARG1 (e/eat-01 :ARG0 i/i)) :op2 (p/permit-01 :ARG1 (\textbf{e2/eat-01} :ARG0 y/you))}
\end{tabular}
}
\end{center}
\caption{Right node raising with shared main verb: ``I should and you may eat''. The derived AMR has a reentrancy for the \SEM{eat-01} predicate where there should be a separate copy of the predicate.}
\label{predicate-ellipsis}
\end{figure}


\end{document}